\documentclass{article} 
\pdfoutput=1
\usepackage[margin=1in]{geometry}
\usepackage[utf8]{inputenc}
\usepackage[T1]{fontenc}
\usepackage{lmodern}

\usepackage{authblk}
\usepackage{cite}
\usepackage{amsmath,amssymb,amsfonts}
\usepackage{graphicx}
\usepackage{subcaption}
\usepackage{booktabs}
\usepackage{textcomp}
\usepackage{algorithm}
\usepackage{algorithmic}
\usepackage{caption}
\usepackage[colorlinks=true,linkcolor=blue,citecolor=blue,urlcolor=blue]{hyperref}
\usepackage{etoolbox}
\AtBeginEnvironment{table}{\footnotesize}
\AtBeginEnvironment{algorithm}{\footnotesize}

\title{Non-Contact Physiological Monitoring in Pediatric Intensive Care Units via Adaptive Masking and Self-Supervised Learning}

\author[1]{Mohamed Khalil Ben Salah}
\author[2]{Philippe Jouvet}
\author[1]{Rita Noumeir}

\affil[1]{Biomedical Information Processing Laboratory, École de Technologie Supérieure, University of Quebec, Montreal, QC, Canada}
\affil[2]{Pediatric Intensive Care Unit, CHU Sainte-Justine, Montreal, QC, Canada}

\date{} 

\begin{document}

\maketitle
\thispagestyle{empty}
\begin{abstract}
Continuous monitoring of vital signs in Pediatric Intensive Care Units (PICUs) is essential for early detection of clinical deterioration and effective clinical decision-making. However, contact-based sensors such as pulse oximeters may cause skin irritation, increase infection risk, and lead to patient discomfort.
Remote photoplethysmography (rPPG) offers a contactless alternative to monitor heart rate using facial video, but remains underutilized in PICUs due to motion artifacts, occlusions, variable lighting, and domain shifts between laboratory and clinical data.
We introduce a self-supervised pretraining framework for rPPG estimation in the PICU setting, based on a progressive curriculum strategy. The approach leverages the VisionMamba architecture and integrates an adaptive masking mechanism, where a lightweight Mamba-based controller assigns spatiotemporal importance scores to guide probabilistic patch sampling. This strategy dynamically increases reconstruction difficulty while preserving physiological relevance.
To address the lack of labeled clinical data, we adopt a teacher–student distillation setup. A supervised expert model, trained on public datasets, provides latent physiological guidance to the student. The curriculum progresses through three stages: clean public videos, synthetic occlusion scenarios, and unlabeled videos from 500 pediatric patients.
Our framework achieves a 42\% reduction in mean absolute error relative to standard masked autoencoders and outperforms PhysFormer by 31\%, reaching a final MAE of 3.2 bpm. Without explicit region-of-interest extraction, the model consistently attends to pulse-rich areas and demonstrates robustness under clinical occlusions and noise.

\end{abstract}

\noindent\textbf{Keywords:}
Remote Photoplethysmography, Self-Supervised Learning, Masked Autoencoder,
Teacher-Student Distillation, Pediatric Intensive Care Unit.

\section{Introduction}
\label{sec:introduction}
Contactless monitoring of vital signs in Pediatric Intensive Care Units (PICUs) is essential to ensure patient safety and support early clinical intervention. Traditional contact-based sensors, such as electrocardiograms (ECGs) and pulse oximeters, are widely used but remain suboptimal for fragile neonates. These sensors can irritate sensitive skin and increase the risk of infection~\cite{bib1}. Remote photoplethysmography (rPPG), which estimates blood volume pulse (BVP) from subtle changes in facial color, provides a non-contact alternative. Since the COVID-19 pandemic, rPPG has attracted growing interest across telemedicine, fitness applications, and clinical monitoring. In the PICU setting, its unobtrusive nature offers the potential to reduce patient discomfort, minimize caregiver handling, and enable continuous monitoring. A reliable rPPG solution that provides heart rate monitoring, could improve pediatric critical care by eliminating the need for adhesive or invasive sensors.

Despite its potential, deploying rPPG models in clinical settings remains challenging. The PICU environment presents multiple sources of signal degradation, including occlusions (e.g., tubes, caregivers), variable lighting, motion artifacts, and high inter-subject variability. Moreover, supervised learning approaches are limited by the scarcity of labeled clinical data, as video acquisition in PICUs is ethically sensitive and logistically constrained. Public datasets such as UBFC-rPPG and PURE are collected under controlled conditions and do not generalize well to real-world clinical environments. This domain mismatch often results in poor performance when models trained on lab data are applied in clinical practice~\cite{bib2}.

To address these challenges, we introduce a self-supervised pretraining framework for rPPG estimation, specifically designed for the PICU context. Our framework is trained on an Institutional Review Board (IRB)-approved dataset of 500 pediatric patient videos and follows a curriculum learning strategy. The training progresses in three stages: initial learning on clean public datasets, followed by synthetic occlusion augmentation, and culminating in large-scale pretraining on real PICU data. This staged approach improves model robustness to clinically relevant artifacts and environmental variability.

The framework is based on a masked autoencoder architecture inspired by VideoMAE~\cite{bib4}, and introduces two core innovations. First, we incorporate an Adaptive Masking Network (AMN) that replaces conventional random masking. The AMN uses a Mamba-based controller trained via policy gradient reinforcement learning to assign importance scores to each spatiotemporal patch. By sampling binary masks that occlude highly informative regions, typically those containing strong pulsatile signals, the model is forced to develop more generalizable and physiologically meaningful representations. This mechanism allows the model to implicitly attend to key facial regions, without requiring explicit region-of-interest (ROI) extraction, and enhances generalization under challenging conditions such as occlusion and noise.

Second, we integrate a multi-objective training scheme to guide learning. The main generative objective combines Mean Squared Error (MSE) with a pixel-level Pearson correlation term to ensure accurate temporal modeling of BVP dynamics. While this promotes global spatiotemporal learning, it does not guarantee alignment with physiological targets. To introduce this inductive bias, we adopt a teacher–student distillation approach. A supervised expert model provides high-quality reference signals, and the student network is trained to align its predictions with these targets. This process reinforces the extraction of clinically meaningful features, even in cases of partial visibility or degraded video quality.

\noindent\textbf{Our key contributions are as follows:}
\begin{itemize}
    \item We introduce a contactless vital-sign monitoring framework for PICUs that lowers skin lesions, and enables continuous physiological monitoring.
    \item We propose a self-supervised rPPG framework that integrates masked autoencoding with Pearson correlation and physiological distillation to learn robust pulsatile dynamics from facial videos.
    \item We present an adaptive masking strategy using a Mamba-based controller, which dynamically selects spatiotemporal patches to improve robustness against occlusions, motion artifacts, and variable lighting.
    \item We validate our method on a large IRB-approved dataset of 500 pediatric patients and demonstrate its effectiveness for rPPG and heart rate estimation under realistic PICU conditions.
\end{itemize}

\section{Related Works}

\subsection{Remote Photoplethysmography (rPPG)}

Conventional rPPG methods typically follow a two-stage pipeline involving facial region detection and signal extraction based on RGB intensity variations that reflect blood volume pulse (BVP) dynamics. Early signal processing techniques such as Independent Component Analysis (ICA)~\cite{bib8}, CHROM~\cite{bib9}, and Plane-Orthogonal-to-Skin (POS)~\cite{bib10} were developed to isolate pulsatile components while mitigating motion and illumination artifacts. Although effective under controlled conditions, these handcrafted approaches rely on fixed assumptions about lighting and skin reflectance, which limit their applicability in clinical environments such as the PICU, where motion, occlusion, and sensor noise are common. These constraints have motivated the shift toward deep learning models that offer improved robustness through data-driven learning.

Initial deep learning approaches for rPPG estimation employed 2D convolutional neural networks (CNNs). Some introduced motion-based attention mechanisms~\cite{bib11}, while others incorporated video magnification~\cite{bib12} or emphasized spatial features linked to pulse signals~\cite{bib13}. Subsequent models, including EfficientPhys~\cite{bib14} and Multi-Task Temporal Shift Convolutional Attention Network (MTTS-CAN)~\cite{bib15}, proposed architectural refinements to better capture temporal dynamics. However, 2D CNNs remain limited in modeling spatiotemporal dependencies. This led to the use of 3D CNNs, such as rPPGNet~\cite{bib16} and PhysNet~\cite{bib17}, which process entire video volumes. Further improvements were introduced through attention-based architectures~\cite{bib18, bib19} designed to enhance signal localization.

Other strategies include hybrid CNN-RNN models~\cite{bib20, bib21, bib22} to model temporal dependencies, and meta-learning methods~\cite{bib23} to improve generalization across subjects and domains. Despite these advances, 3D CNNs often remain more effective in clinical applications due to their ability to jointly model spatial and temporal features.

More recently, transformer-based models have been adopted for rPPG estimation~\cite{bib24}. Vision Transformers (ViT)~\cite{bib25} inspired specialized designs such as TransPPG~\cite{bib26} and PhysFormer~\cite{bib28}, which incorporate motion awareness and periodicity constraints. TimeSformer~\cite{bib29} introduced divided space-time attention, and rPPGTR~\cite{bib30} extended this to combine local CNN features with global attention for robust signal reconstruction. However, these approaches still depend on supervised training and large-scale labeled datasets, which are often unavailable in clinical contexts.

Our work builds on these developments by proposing a self-supervised learning framework that leverages large-scale unlabeled PICU data. The model is trained using a staged curriculum designed to capture the variability of real-world clinical conditions without requiring manual annotations.

\subsection{Self-Supervised Learning for Video Understanding}

Self-supervised learning (SSL) has emerged as a promising direction for learning video representations from unlabeled data, particularly in medical applications where annotation is both costly and privacy-sensitive~\cite{bib3}. Pretext tasks such as contrastive learning~\cite{Chen2020SimCLR, grill2020bootstrap, Caron2021EmergingProperties} and masked reconstruction~\cite{He2022MAE} have demonstrated strong performance in visual domains.

Among these, generative approaches based on Masked Image Modeling (MIM) have shown particular effectiveness. The Masked Autoencoder (MAE)~\cite{He2022MAE} employs an asymmetric encoder–decoder structure to reconstruct masked patches, enabling efficient learning of semantically meaningful features. This approach has been extended to video with VideoMAE~\cite{bib4}, which applies high masking ratios (90–95\%) across spatiotemporal tokens using a tube-based masking pattern. By leveraging temporal redundancy, VideoMAE achieves robust pretraining while maintaining computational efficiency through a lightweight Vision Transformer (ViT) encoder. This design highlights the importance of pretraining directly on domain-specific video datasets to improve downstream task performance.

Recent extensions such as Unmasked Teacher (UMT)~\cite{bib6} build on this paradigm through a teacher–student framework that guides the masking process. Instead of randomly selecting patches, UMT uses teacher-derived attention maps to mask less informative regions, thereby improving semantic alignment and enhancing representation learning efficiency.

Despite these advances, most existing SSL methods rely on static or random masking strategies that do not adapt to the content variability in clinical videos. In tasks such as rPPG estimation, spatial heterogeneity caused by motion, occlusion, and physiological signal variability requires a more targeted masking approach. To address this, we propose an adaptive masking mechanism driven by a lightweight Mamba-based policy network. Unlike fixed masking schemes, our method learns to identify and suppress visually  physiologically regions, guiding the reconstruction process toward clinically informative areas. This improves feature learning under the complex and variable conditions characteristic of the PICU environment.

\subsection{SSL for Remote Photoplethysmography}

SSL has emerged as an effective strategy to address the scarcity and high acquisition cost of physiological annotations in remote photoplethysmography. Recent SSL frameworks have explored diverse objectives to improve generalization under conditions involving motion, occlusion, illumination changes, and noise~\cite{Hasan2022SelfrPPG, Zhang2024SSLrPPG, Wang2021Spatiotemporal, Xiao2024SimFuPulse}.

Several approaches have focused on contrastive learning, where the objective is to distinguish between temporally or spatially aligned positive pairs and unaligned negatives. Gideon et al.~\cite{Gideon2021Contrastive} introduced a fully self-supervised method that relies on weak priors in the frequency and temporal domains, removing the need for labels. Contrast-Phys~\cite{Sun2022ContrastPhys} and its extension Contrast-Phys+~\cite{Sun2024ContrastPhysPlus} applied 3D convolutional backbones with contrastive objectives to extract physiological representations from facial videos. These models achieved improved robustness to motion and noise, performing on par with supervised baselines.

In parallel, generative SSL techniques based on masked autoencoding have demonstrated advantages in learning structural and periodic signal information. rPPG-MAE~\cite{Liu2023rPPGMAE} used a masked reconstruction objective to model the self-similarity of BVP signals, showing improved resilience to motion and illumination variability. Notably, the authors emphasized that pretraining performance depends more on dataset quality than size. Yue et al.~\cite{Yue2023SSLrPPG} proposed a frequency-aware SSL method that combines spatial and frequency domain augmentations with a learnable frequency transform module, enabling better encoding of signal periodicity in the absence of labels.

SSL methods incorporating teacher–student structures or pseudo-labeling have also gained traction. PhySU-Net employed pseudo-labels derived from conventional signal extraction pipelines to guide a masked image reconstruction task. Similarly, Li et al.~\cite{Li2023PseudoLabels} proposed a co-rectification strategy to improve training stability in the presence of noisy pseudo-supervision. These works aim to combine handcrafted signal priors with learned representations.

Transformer-based SSL models have recently gained momentum for capturing long-range spatiotemporal dependencies. RS-rPPG~\cite{Savic2024RSrPPG} leveraged temporal augmentation and a transformer backbone, achieving improved performance over earlier contrastive models. ST-Phys~\cite{Cao2024STPhys} demonstrated strong robustness to occlusion and noise using a lightweight architecture. SimPPG~\cite{Bhattachrjee2023SimPPG} introduced a region-based non-contrastive framework using positive pairs from the same subject, allowing effective physiological representation learning without explicit labels. TransPhys~\cite{Wang2023TransPhys} further advanced this line of work by applying contrastive transformer pretraining for global feature learning, outperforming existing SSL benchmarks in rPPG estimation.

These studies highlight the growing potential of SSL for rPPG under uncontrolled conditions. However, most prior work relies on uniform masking or augmentation schemes that ignore the spatial and temporal variability present in clinical data. Such models typically treat all input regions as equally informative, lacking mechanisms to focus on physiologically relevant areas. In addition, current frameworks often omit structured curricula or physiological guidance, limiting their capacity to generalize to the complex visual dynamics of PICU recordings.

To address these challenges, we introduce a unified self-supervised framework that integrates multiple components previously studied in isolation. Specifically, we propose an adaptive masking strategy driven by a lightweight Mamba-based controller, a teacher–student distillation mechanism that transfers physiological priors from a supervised model, and a progressive curriculum learning schedule that bridges the gap between controlled datasets and real-world PICU data. This combination improves model robustness to occlusions, motion artifacts, lighting variability, and limited annotation, which are characteristic of intensive care settings.

\subsection{State Space Models (SSMs)}

Transformers~\cite{bib24} have become the dominant architecture for sequential data modeling in natural language processing and computer vision due to their ability to capture global dependencies through multi-head self-attention. However, their quadratic complexity with respect to sequence length presents scalability challenges, particularly for high-resolution images and long video sequences.

State Space Models (SSMs) have emerged as a scalable alternative for long-sequence modeling, offering linear computational complexity. Structured State Space sequence models (S4)~\cite{Gu2021StateSpaces} demonstrated that reparameterized state space representations can capture long-range dependencies efficiently. S4 introduced low-rank updates and diagonal stabilization by normalizing parameter matrices. Subsequent variants, including S5~\cite{Smith2023SimplifiedSSL}, H3~\cite{Fu2023Hippos}, and GSS~\cite{Mehta2022GSS}, have explored trade-offs in stability, expressivity, and computational efficiency.

Mamba~\cite{Mamba} recently introduced a data-dependent SSM mechanism that enables selective processing of input sequences. By making transition matrices input-dependent and optimizing for parallel execution, Mamba achieves efficient long-range modeling with linear complexity. It has demonstrated strong performance across large-scale datasets and has been applied to diverse tasks~\cite{Pioro2024MoE-Mamba}.

In the visual domain, Mamba has been adapted for spatial and spatiotemporal modeling. Vision Mamba (ViM)~\cite{VisionMamba} extends Mamba to 2D using bidirectional scans for improved spatial representation. VMamba~\cite{VMamba} introduces a four-directional Selective Scan (SS2D) for enhanced contextual propagation. EfficientVMamba~\cite{Pei2024EfficientVMamba} reduces computational cost via atrous sampling, while LocalMamba~\cite{Huang2024LocalMamba} focuses on local context through windowed scanning and spatial-channel attention. Mamba-based backbones have also been adopted in medical image analysis, including segmentation and detection tasks~\cite{Ruan2024VM-UNet, Liu2024SwinUMamba}.

These developments show the potential of SSM-based models to process long visual sequences with improved scalability. In this work, we leverage Vision Mamba to model clinical video sequences, where capturing temporally extended patterns and ensuring computational efficiency are critical.

\section{Methodology}
\label{sec:method}

\subsection{Overview}
\label{ssec:overview}

\begin{figure}[t]
    \centering
    \includegraphics[width=0.8\textwidth]{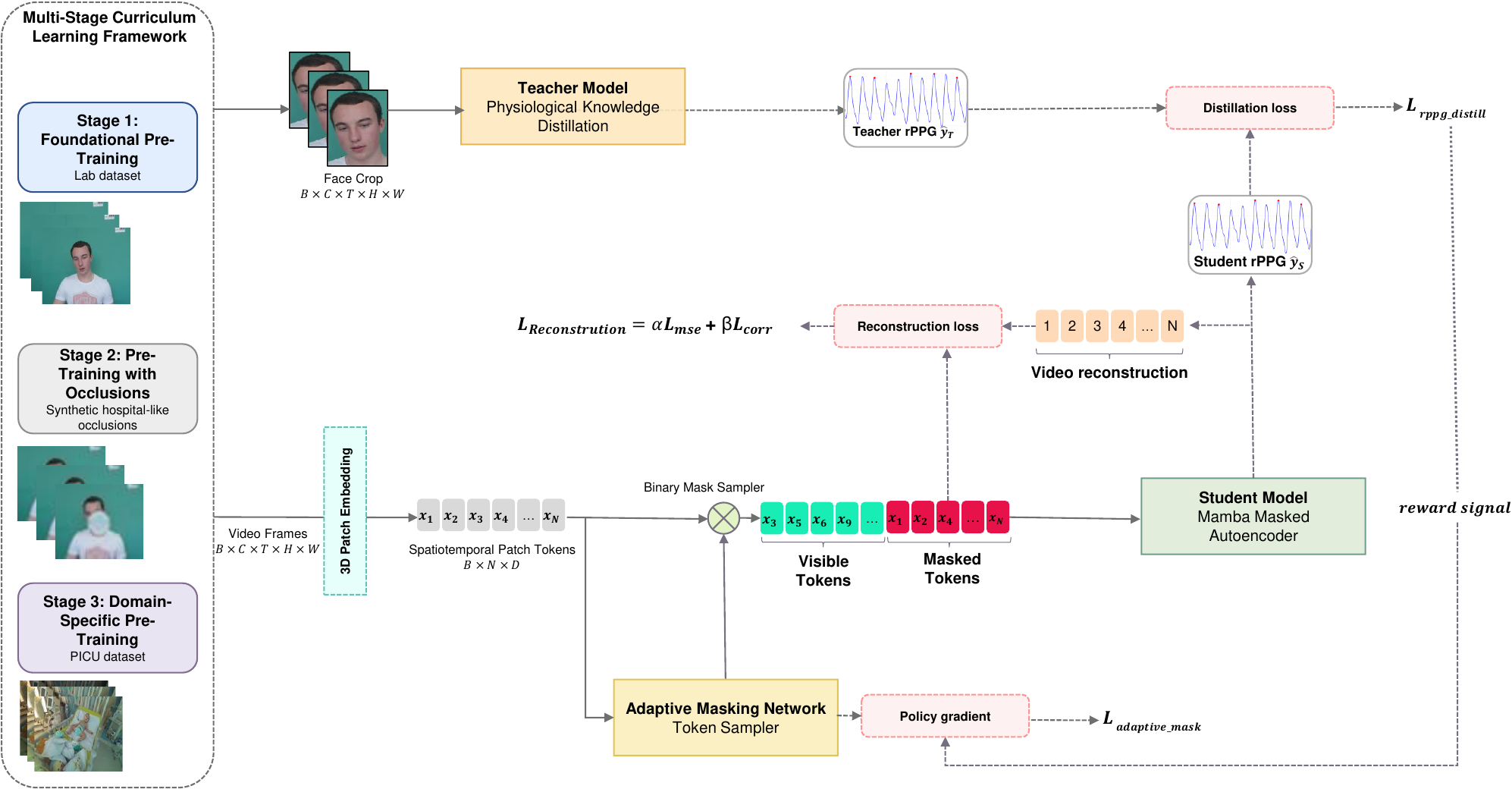} 
    \caption{Overview of our curriculum-based framework for robust rPPG estimation. The student model learns by reconstructing masked patches and predicting the rPPG signal from visible tokens. Training is guided by an expert PhysMamba teacher model and a learnable Adaptive Masking Network (AMN) optimized via policy gradient reinforcement.}
    \label{fig:framework}
\end{figure}

This work proposes a self-supervised framework for robust remote photoplethysmography estimation in Pediatric Intensive Care Unit environments. The method combines adaptive masked autoencoding with physiological knowledge distillation to improve signal robustness under common challenges such as occlusion, lighting variation, and patient motion.

As shown in Figure~\ref{fig:framework}, the architecture includes three components: a student model, a teacher model, and a learnable Adaptive Masking Network (AMN).

The student model is optimized on two tasks. It receives spatiotemporal tokens from an input video, a subset of which is masked by the AMN. The model must: (i) reconstruct the masked patches via a lightweight decoder in a self-supervised setting, and (ii) estimate the rPPG signal from the unmasked tokens through a regression head. The PhysMamba teacher provides the reference rPPG signal, enabling physiological knowledge transfer.

The AMN is the core novelty of this framework. Rather than using random masking, it learns to select and mask informative spatiotemporal tokens based on their importance. It consists of lightweight Mamba blocks that assign importance scores to tokens and apply differentiable Gumbel-Top-K sampling to choose which tokens to mask. AMN is trained via a policy gradient strategy using the student’s rPPG loss as a reward. The masking policy is rewarded when it increases the student's prediction error, pushing the student to extract more robust and generalizable features.

To improve generalization, we adopt a three-stage curriculum. The model is first trained on clean laboratory videos, then on synthetically occluded samples, and finally on real PICU data from 500 pediatric patients.

\subsection{VisionMamba Student Model}
As illustrated in Figure~\ref{fig:amn}, the student architecture consists of three main components: a patch tokenizer, an encoder, and a decoder head.

\subsubsection{Patch Tokenizer}

The student model processes an input video \(\mathbf{X} \in \mathbb{R}^{C \times T \times H \times W}\), where \(C\) is the number of channels, \(T\) the number of frames, and \(H \times W\) the spatial resolution. A 3D convolutional patch embedding layer with kernel size and stride \((t, h, w)\) partitions the input into non-overlapping tubelets, generating patch tokens \(\mathbf{X}_p \in \mathbb{R}^{N \times D}\), where \(N\) is the number of tokens and \(D\) the embedding dimension. To retain spatial and temporal ordering, fixed sinusoidal positional embeddings are added to the patch tokens.

\subsubsection{Encoder Block}
The Adaptive Masking Network generates a binary mask that retains a subset of visible tokens \(\mathbf{X}_{\text{vis}} \in \mathbb{R}^{K \times D}\), where \(K < N\). These visible tokens are passed through an encoder composed of \(L\) stacked bidirectional Mamba blocks with residual connections, followed by a normalization layer. The output is linearly projected to match the decoder's input dimension and combined with learnable mask tokens corresponding to the masked positions. This combined sequence is used for reconstruction in the decoder.

\subsubsection{Decoder Head}
The decoder receives the visible token representations, projected into the decoder embedding dimension, along with the learnable mask tokens corresponding to the masked positions. Fixed positional embeddings are added to both sets, which are then concatenated to form the full decoder input. This sequence is processed by a stack of \(D\) Mamba blocks, followed by a normalization layer. The decoder head, implemented as a linear projection, reconstructs only the masked tokens, yielding patch-level outputs for self-supervised learning.

In parallel, the model includes an rPPG prediction head. It aggregates both visible and masked token embeddings using mean pooling and feeds the result into a multilayer perceptron (MLP) with dropout and non-linear activation to predict the rPPG waveform of length \(T\). This dual objective encourages the model to learn both spatial reconstruction and physiologically relevant temporal dynamics.

\subsection{Adaptive Masking Network}
Standard masked autoencoders typically apply static or random masking schemes that lack content awareness, often removing physiologically informative regions or retaining uninformative ones. To address this limitation, we introduce an Adaptive Masking Network, a learnable policy module that selectively masks the most informative spatiotemporal tokens in an adversarial manner. By exposing the student only to low-signal or ambiguous tokens, the AMN increases task difficulty and encourages the extraction of robust features relevant to physiological dynamics.

The AMN operates on the patch embeddings \(\mathbf{x} \in \mathbb{R}^{B \times N \times D}\), derived from a 3D convolutional projection of the input video \(\mathbf{x}_{\text{video}} \in \mathbb{R}^{B \times C \times T \times H \times W}\). Instead of relying on attention-based architectures, the AMN leverages a lightweight stack of Mamba blocks to efficiently capture long-range spatial and temporal dependencies. The output is passed through a shallow MLP to compute per-token importance scores.

\begin{figure}[t]
  \centering
  \includegraphics[width=0.8\textwidth]{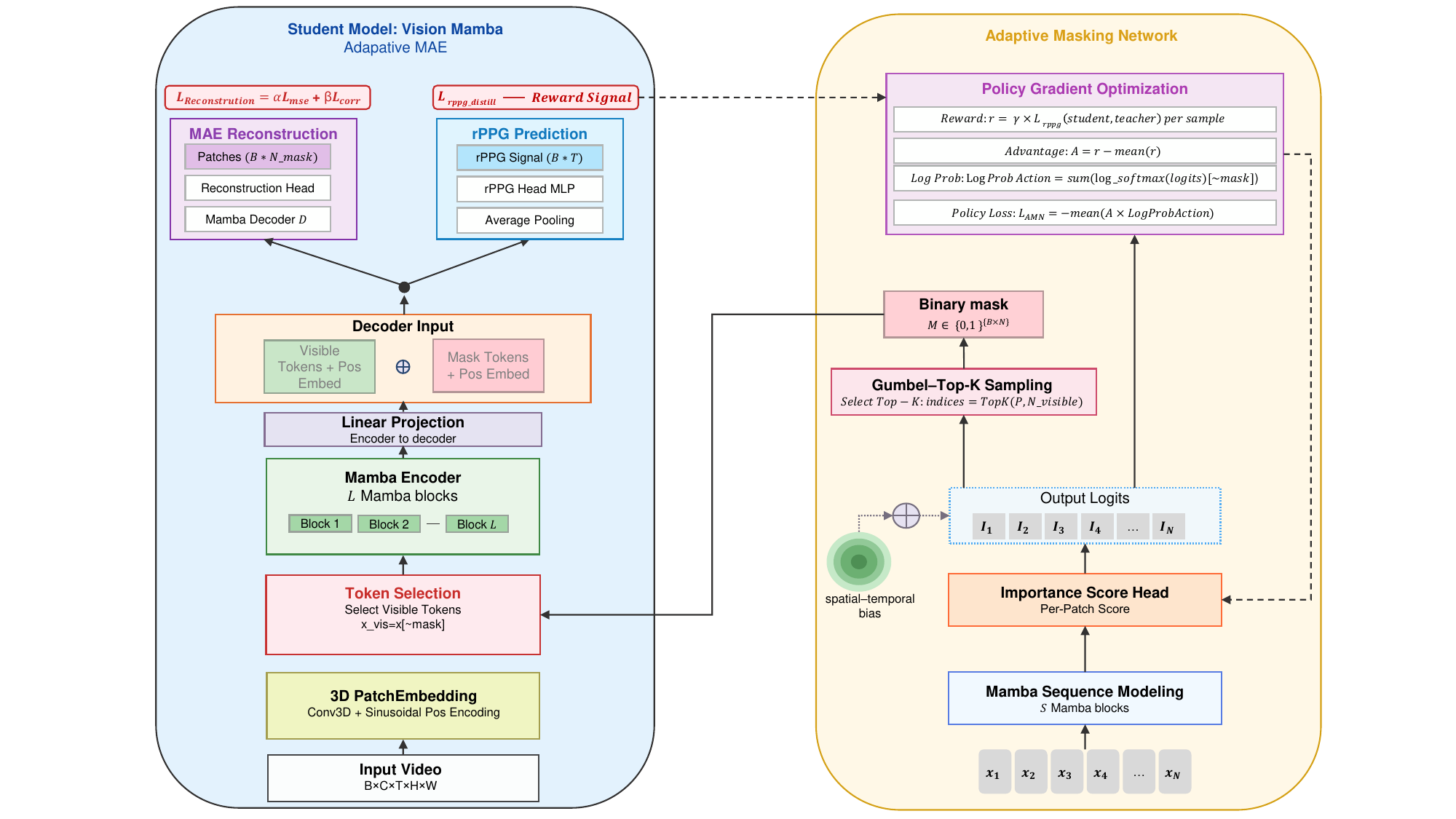}
  \caption{Detailed architecture of the Adaptive Masking Network (AMN). The AMN computes token importance scores using Mamba blocks and selects a visible subset via Gumbel--Top-K sampling. Policy gradient optimization uses the rPPG distillation loss as reward to update the AMN. The student reconstructs the masked tokens and predicts the rPPG waveform.}
  \label{fig:amn}
\end{figure}

To guide masking toward relevant regions, a spatial prior bias is added, favoring central facial areas. Gumbel noise is introduced to ensure sampling variability, and a Top-K operation selects the top \((1 - r)\%\) tokens, where \(r\) is the target masking ratio. The remaining tokens are masked. The final visibility mask is binary and differentiable, as illustrated in Figure~\ref{fig:amn}.

This mask is applied to the token sequence, allowing the student to process only the retained tokens, while the masked positions are replaced with a learned embedding. The AMN is optimized not by conventional backpropagation but via policy gradient reinforcement learning. The reward signal is defined by the student’s rPPG prediction loss, with higher errors corresponding to more challenging masks. This setup drives the AMN to occlude high-signal regions, such as the forehead or cheeks, forcing the student to learn from sparse and degraded observations. The full procedure for generating the mask is described in Algorithm~\ref{alg:amn}.

\begin{table*}[htbp]
  \centering
  \scriptsize 
  
  \begin{minipage}[t]{0.48\textwidth}
    \begin{algorithm}[H] 
    \footnotesize
      \caption{Adaptive Masking Network (AMN)}
      \label{alg:amn}
      \begin{algorithmic}[1]
        \REQUIRE $\mathbf{X}$, parameters $\theta$, $\tau$, ratio $\rho$
        \ENSURE Mask $\mathbf{M}$, scores $\mathbf{S}$
        
        \STATE \textbf{Init:} $N_{vis} \leftarrow \lfloor N \times (1 - \rho) \rfloor$; Init $\mathbf{B}_{spatial}$
        \STATE \textbf{Stage 1: Patch Embedding}
        \STATE $\mathbf{P} \leftarrow \text{PatchEmbed}(\mathbf{X}) + \text{PosEnc}$
        
        \STATE \textbf{Stage 2: Importance Computation}
        \STATE $\mathbf{H} \leftarrow \mathbf{P}$
        \FOR{each MambaBlock in AMN}
            \STATE $\mathbf{H} \leftarrow \text{MambaBlock}(\mathbf{H})$
        \ENDFOR
        \STATE $\mathbf{S} \leftarrow \text{Squeeze}(\text{ImpHead}(\text{LN}(\mathbf{H}))) + \mathbf{B}_{spatial}$
        
        \STATE \textbf{Stage 3: Differentiable Sampling}
        \STATE $\mathbf{S}_{logits} \leftarrow \text{Clamp}(\mathbf{S} / \tau, -10, 10)$
        \STATE $\mathbf{G} \leftarrow \text{GumbelNoise}(\text{Uniform}(0,1))$
        \STATE $idx_{vis} \leftarrow \text{TopK}(\mathbf{S}_{logits} + \mathbf{G}, N_{vis})$
        \STATE $\mathbf{M} \leftarrow \mathbf{1}$; $\mathbf{M}[idx_{vis}] \leftarrow 0$
        
        \RETURN $\mathbf{M}, \mathbf{S}_{logits}$
      \end{algorithmic}
    \end{algorithm}
  \end{minipage}
  \hfill 
  \begin{minipage}[t]{0.48\textwidth}
    \begin{algorithm}[H] 
    \footnotesize
      \caption{Policy Gradient Training}
      \label{alg:policy_gradient}
      \begin{algorithmic}[1]
        \REQUIRE Student $f_s$, Teacher $f_t$, AMN $g_\theta$, $\beta, \alpha$
        \ENSURE Updated $\theta$
        
        \STATE \textbf{Forward Pass:}
        \STATE $\mathbf{M}, \mathbf{S} \leftarrow g_\theta(\mathbf{X})$ \COMMENT{Gen adaptive mask}
        \STATE $\mathbf{X}_{masked} \leftarrow \text{ApplyMask}(\mathbf{X}, \mathbf{M})$
        \STATE $\hat{\mathbf{y}}_{s} \leftarrow f_s(\mathbf{X}_{masked})$; $\mathbf{y}_{t} \leftarrow f_t(\mathbf{X}_{face})$
        
        \STATE \textbf{Reward Computation:}
        \STATE $\mathcal{L}_{distill} \leftarrow 1 - \rho(\hat{\mathbf{y}}_{s}, \mathbf{y}_{t})$
        \STATE $R \leftarrow \mathcal{L}_{distill}$; $b \leftarrow \text{Mean}(R)$
        \STATE $A \leftarrow (R - b) \times \beta$ \COMMENT{Advantage}
        
        \STATE \textbf{Policy Gradient Loss:}
        \STATE $\pi(\mathbf{M}|\mathbf{S}) \leftarrow \text{LogSoftmax}(\mathbf{S})$
        \STATE $\log p(\mathbf{M}) \leftarrow \sum_{i \in \neg\mathbf{M}} \pi_i$
        \STATE $\mathcal{L}_{adaptive} \leftarrow -\text{Mean}(\log p(\mathbf{M}) \cdot \text{det}(A)) \cdot \alpha$
        
        \STATE \textbf{Update:} $\theta \leftarrow \theta - \eta_a \nabla_\theta \mathcal{L}_{adaptive}$
        
        \RETURN $\theta$
      \end{algorithmic}
    \end{algorithm}
  \end{minipage}
\end{table*}

Formally, given the importance logits \(\mathbf{l} \in \mathbb{R}^{B \times N}\) and the sampled binary mask \(\mathbf{m} \in \{0,1\}^{B \times N}\), the log-probability of the selected masking action is computed from the softmax distribution over \(\mathbf{l}\). The policy gradient loss is defined as:

\begin{equation}
\mathcal{L}_{\text{PG}} = -\mathbb{E} \left[ \left( \mathcal{L}_{\text{rPPG}} - \bar{\mathcal{L}} \right) \cdot \log p(\mathbf{m}) \right]
\end{equation}

where \(\mathcal{L}_{\text{rPPG}}\) is the signal-level distillation loss between the student and teacher rPPG waveforms, \(\bar{\mathcal{L}}\) is a batch-wise baseline for variance reduction, and \(p(\mathbf{m})\) is the probability of the sampled visible tokens.

The AMN is jointly trained with the student model but does not receive direct gradients from the decoder or the regression head. Instead, it learns indirectly through its effect on the student's performance, effectively acting as a hard instance generator that strengthens feature robustness.

\subsection{Physiological Knowledge Distillation}
\label{ssec:distillation}
Although self-supervised masked reconstruction supports general representation learning from unlabeled clinical videos, it lacks explicit physiological guidance, which is critical for capturing the fine-grained temporal dynamics required for accurate rPPG estimation. To overcome this limitation, we integrate a distillation mechanism in which a pre-trained expert model, PhysMamba~\cite{bib7}, serves as a fixed teacher during pretraining.

PhysMamba is a supervised rPPG model trained on clean, face-cropped clean datasets. It processes DiffNormalized frames, where each frame is temporally normalized relative to its neighbors to enhance pulsatile information and suppress motion and lighting variability. Within our framework, PhysMamba provides physiological supervision by generating high-fidelity reference signals, which guide the student during masked autoencoding. While knowledge distillation is commonly used in general vision tasks, its use with a domain-specific Mamba model for self-supervised rPPG pretraining in clinical video remains largely unexplored. This design embeds physiological priors into the learned representations and improves temporal consistency in the presence of clinical artifacts.

The student incorporates a dedicated rPPG prediction head, which regresses the waveform from pooled token embeddings. Specifically, global average pooling is applied across both visible and masked spatiotemporal tokens before decoding. This provides the rPPG head with access to the overall temporal context required for signal reconstruction.

To enforce alignment between the student’s predictions and the teacher-generated reference signals, we apply a signal-level distillation loss based on negative Pearson correlation:

\begin{equation}
\mathcal{L}_{\text{distill}} = 1 - \rho(\hat{\mathbf{y}}_{\text{student}}, \mathbf{y}_{\text{teacher}})
\end{equation}

where \(\rho(\cdot, \cdot)\) denotes the Pearson correlation coefficient between the predicted and teacher-generated rPPG waveforms. No feature-level alignment is enforced, avoiding over-constraining the latent space and ensuring that learning remains focused on waveform morphology and temporal coherence.

This loss serves a dual purpose: it provides a direct physiological supervision signal and simultaneously functions as a reward for the Adaptive Masking Network. Coupling the distillation loss with the masking policy gradient encourages the AMN to suppress informative patches, thereby increasing task difficulty and reinforcing robust temporal learning in the student.

Through this distillation mechanism, the student acquires the ability to infer physiologically accurate waveforms even under severe occlusion or partial visibility, mirroring realistic challenges encountered in PICU settings.

\subsection{Training Objectives and Optimization}
\label{sec:training_objectives}
The training framework jointly optimizes three complementary objectives that promote robust learning of physiological features. The total loss combines masked reconstruction, physiological signal distillation, and masking policy optimization through a weighted formulation. To decouple feature learning from masking behavior, we adopt a dual-optimizer setup.

\subsubsection{Masked Reconstruction Objective}

The reconstruction loss focuses exclusively on masked tokens and consists of two components designed to preserve both spatial fidelity and temporal coherence. Let $\mathbf{X} \in \mathbb{R}^{B \times C \times T \times H \times W}$ denote the input video and $\mathbf{M} \in \{0,1\}^{B \times N}$ the binary mask generated by AMN, where $N$ is the number of spatiotemporal patches. The overall reconstruction loss is:

\begin{equation}
    \mathcal{L}_{\text{recon}} = \mathcal{L}_{\text{pixel}} + \mathcal{L}_{\text{corr}}.
\end{equation}

\paragraph{Pixel-Level MSE}
This term penalizes local reconstruction errors by minimizing per-pixel differences over masked patches:

\begin{equation}
    \mathcal{L}_{\text{pixel}} = \frac{1}{|\mathcal{M}|} \sum_{i \in \mathcal{M}} \| \hat{\mathbf{x}}_i - \mathbf{x}_i \|_2^2,
\end{equation}

where $\mathcal{M}$ denotes the masked indices, and $\hat{\mathbf{x}}_i$ and $\mathbf{x}_i$ are the reconstructed and ground-truth patches, respectively.

Minimizing pixel-wise error alone is insufficient for rPPG estimation. It often leads to temporally static outputs that ignore the subtle brightness oscillations encoding physiological signals.

\paragraph{Global Pearson Correlation}

To address this limitation, we introduce a correlation-based term that captures global temporal dynamics across masked regions. The correlation is computed over flattened patch sequences:

\begin{equation}
    \mathcal{L}_{\text{corr}} = 1 - \frac{1}{B} \sum_{i=1}^{B} \rho_i,
\end{equation}

where $\rho_i$ is the Pearson correlation between predicted and ground-truth sequences for sample $i$.

This loss integrates several physiological priors essential for waveform reconstruction. First, temporal coherence is enforced by ensuring masked patches span the full duration \(T\), covering multiple cardiac cycles. This allows the model to preserve phase-consistent oscillations caused by pulsatile blood flow. Second, spatial consistency is encouraged, reflecting the fact that facial perfusion induces coordinated brightness changes across multiple regions such as the cheeks, nasal bridge, and forehead. Finally, the correlation term promotes preservation of dynamic range. Pearson correlation is insensitive to amplitude but penalizes shape mismatch, discouraging flat or low-variance reconstructions even when MSE is minimized. These constraints guide the model toward physiologically meaningful predictions.

\subsubsection{Physiological Distillation Objective}
To embed physiological priors during pretraining, we apply knowledge distillation from a fixed teacher model trained on clean, face-cropped videos. The teacher outputs reference rPPG signals from DiffNormalized input. The student predicts an rPPG waveform from globally pooled token embeddings. The distillation loss is computed as:

\begin{equation}
    \mathcal{L}_{\text{distill}} = 1 - \rho(\hat{\mathbf{y}}_s, \mathbf{y}_t),
\end{equation}

where $\hat{\mathbf{y}}_s$ and $\mathbf{y}_t$ are the normalized student and teacher signals, respectively. This term aligns temporal dynamics and encourages physiologically plausible predictions without restricting intermediate features.

\subsubsection{Policy Gradient for Adaptive Masking}

The Adaptive Masking Network is trained to suppress signal-rich patches, making the reconstruction and regression tasks more challenging. Given the per-token importance scores $\mathbf{l} \in \mathbb{R}^{B \times N}$, we apply differentiable Top-$K$ sampling by perturbing these scores with Gumbel noise~\cite{Jang2016GumbelSoftmax}:

\begin{equation}
    \tilde{\mathbf{l}}_i = \frac{\mathbf{l}_i}{\tau} + \mathbf{g}_i, \quad \mathbf{g}_i \sim \text{Gumbel}(0, 1),
\end{equation}
where $\tau$ denotes the temperature. The top $(1-r) \times N$ tokens with the highest scores are selected as visible patches. This approach follows the Gumbel–Softmax trick, which enables a continuous relaxation of the Top-$K$ operation and allows gradient flow through the sampling step. As a result, the network can learn content-aware masking strategies while maintaining stochasticity during training.

The AMN is optimized using a policy gradient formulation, where the student’s rPPG distillation loss serves as a task-specific reward. The policy loss is defined as:

\begin{equation}
    \mathcal{L}_{\text{PG}} = -\mathbb{E}_{\mathcal{V}} \left[ A(\mathcal{V}) \cdot \sum_{j \in \mathcal{V}} \log p_j \right],
\end{equation}
where $p_j = \text{softmax}(\mathbf{l})_j$ is the selection probability of token $j$, and the advantage function is:

\begin{equation}
    A(\mathcal{V}) = w_{\text{rppg}} \cdot (\mathcal{L}_{\text{distill}} - b),
\end{equation}

with $w_{\text{rppg}} = 2.0$ and $b$ a baseline computed per batch to reduce variance. This training setup encourages the AMN to discover masking patterns that degrade the student’s signal prediction performance, thereby forcing the model to learn more generalizable and temporally consistent representations. A complete summary of the AMN update procedure is provided in Algorithm~\ref{alg:policy_gradient}.

\subsubsection{Optimization Strategy}
To ensure stable convergence and disentangle representation learning from policy learning, we employ two independent AdamW optimizers. The parameters of the student model, denoted by $\theta_s$, are updated according to:
\begin{equation}
    \theta_s^{(t+1)} = \theta_s^{(t)} - \eta_s \nabla_{\theta_s} (\lambda_{\text{mae}} \mathcal{L}_{\text{recon}} + \lambda_{\text{dist}} \mathcal{L}_{\text{distill}}),
\end{equation}
where the learning rate $\eta_s$ is set to \(10^{-4}\), the weight decay coefficient is \(\lambda_w = 0.05\), and gradients are clipped such that \(\|\mathbf{g}\|_2 \leq 1.0\) to prevent instability during backpropagation.

The parameters of AMN, denoted by \(\phi\), are optimized separately using the policy gradient loss:
\begin{equation}
    \phi^{(t+1)} = \phi^{(t)} - \eta_{\text{amn}} \nabla_{\phi} \mathcal{L}_{\text{PG}},
\end{equation}

with a learning rate \(\eta_{\text{amn}} = 10^{-5}\). Gradients from \(\mathcal{L}_{\text{PG}}\) do not propagate through the student network. This design maintains a clear separation between the learning of visual representations and the adaptation of masking behavior.

\section{Experimental Protocol}
\subsection{Datasets}

\subsubsection{Public Datasets for Pretraining}

We adopt a curriculum-based pretraining strategy using three publicly available rPPG datasets before transitioning to the clinical PICU domain. These datasets introduce increasing variability in lighting, motion, and acquisition setup, enabling the model to progressively learn robustness across controlled and dynamic scenarios.

\begin{itemize}
    \item \textbf{UBFC-rPPG}~\cite{b5}: consists of 42 videos from 42 subjects recorded under indoor controlled conditions using a Logitech C920 HD Pro webcam. Each video is approximately 90 seconds long and recorded at 30~fps with a resolution of $640 \times 480$ pixels. Participants remained relatively still while engaging in a time-constrained mental arithmetic task to induce natural heart rate variation. Ground truth PPG signals were acquired using a CMS50E pulse oximeter sampled at 60~Hz.

    \item \textbf{VIPL-HR}~\cite{niu2018viplhr}: includes 2,378 RGB videos from 107 subjects captured under varying illumination, spontaneous head movements (e.g., talking, gaze shifts), and with multiple acquisition devices (Logitech C310, Intel RealSense F200, Huawei P9), resulting in heterogeneous resolutions and frame rates (25–30~fps). Some sessions were conducted post-exercise to diversify heart rate ranges. Ground truth BVP signals were collected via a CONTEC CMS60C sensor. The dataset introduces variability that facilitates generalization to out-of-distribution domains.

    \item \textbf{ECG-Fitness}~\cite{a5}: contains 204 videos from 17 subjects (14 male, 3 female; aged 20 to 53 years) performing various physical activities such as talking, rowing, and exercising on ellipticals or bikes. Videos were recorded at 30~fps and $1280 \times 720$ resolution under three lighting setups: natural light, 400~W halogen, and 30~W LED. ECG signals were synchronously recorded at 256~Hz. The presence of substantial motion artifacts makes this dataset critical for pretraining in dynamic conditions relevant to clinical scenarios.
\end{itemize}

\subsubsection{Clinical Dataset: The CHU Sainte-Justine PICU Collection}

For domain adaptation and final evaluation, we use a large-scale clinical dataset acquired at the Pediatric Intensive Care Unit of CHU Sainte-Justine (CHU-SJ)~\cite{boivin2023multimodality}. Data collection was approved by the institutional ethics committee (protocols \#2019-2035 and \#2016-1242), with informed parental consent obtained prior to recording. The dataset comprises 30-second video segments from 500 pediatric patients across a broad age range (1 month to 18 years), representing diverse ethnic backgrounds and skin tones.

Recordings were performed using a multimodal acquisition system centered around the Microsoft Azure Kinect DK, capturing high-resolution RGB video at $3840 \times 2160$ resolution and 30~fps. Synchronized physiological signals including PPG, ECG, and oxygen saturation (SpO\textsubscript{2}) were simultaneously collected from Philips IntelliVue MX800 bedside monitors at 125~Hz.

This dataset presents considerable challenges due to the complexity of real-world clinical settings. Videos frequently include occlusions, such as caregiver interventions and medical devices including oxygen masks (6.0\%), ventilator tubes (3.2\%), blankets (5.8\%), and hats (4.4\%). The cohort includes both spontaneously breathing and mechanically ventilated patients, with varying levels of supplemental oxygen. These characteristics provide a rigorous benchmark for evaluating the robustness and deployment feasibility of rPPG models in hospital environments.

\subsection{Training Details}
All experiments were implemented in PyTorch and executed on a server with four NVIDIA A100 GPUs.

\subsubsection{Data Preprocessing and Input Configuration}
For the teacher model, facial regions were detected using YOLO5Face. Bounding boxes were enlarged by a factor of 1.2 to encompass the full face. Each frame was resized to $224 \times 224$ and normalized using ImageNet statistics. Every training sample includes $T = 128$ consecutive frames, equivalent to approximately 4.3 seconds of video at 30~fps. During training, heart rate was estimated via FFT peak detection within the 0.7–3.0~Hz physiological band. For evaluation on 30-second clips, a sliding window of 128 frames was used, with HR computed per window. Window-level HR values were smoothed using a median filter (kernel size 5) and averaged across windows to obtain one estimate per video. Ground-truth PPG was downsampled to 30~Hz and temporally aligned with the video using synchronized timestamps.

The student model uses a tokenizer with a tubelet size of $(t, h, w) = (2, 16, 16)$, producing a sequence of $N = 12{,}544$ spatiotemporal tokens per clip.

\subsubsection{Hospital Occlusion Simulation}

Stage 2 incorporates a Hospital Occlusion Simulator replicating seven common occlusion types observed in the PICU. These include: (1) medical tubes as Bézier curves (thickness 3–8 pixels), (2) oxygen masks as ellipses covering 70\% of the face width, (3) medical tape as rotated rectangles (10–25~$\times$~15–40 pixels), (4) hands as irregular ellipses entering from frame edges, (5) shadows via gradient overlays (intensity 0.3 to 0.7), (6) equipment obstructing 15–30\% of the frame width, and (7) blanket edges modeled using cubic splines.

Each occlusion persists for 20–80\% of the clip with 70\% temporal consistency. Following the curriculum strategy, occlusion probability increases linearly from 0\% to 50\% between epochs 50 and 150, and spatial coverage increases from 10\% to 40\% of the face.

Colors match typical clinical appearances: grayscale (100–180) for equipment, high-intensity white (200–255) for tubes, and skin-tone or blue for gloves and hands. This augmentation increases robustness to visual disruptions encountered in clinical practice, supporting domain adaptation in Stage 3.

\subsubsection{Model and Optimization Hyperparameters}

The VisionMamba student encoder consists of 12 bidirectional Mamba blocks (dim 768). The decoder includes 8 Mamba blocks (dim 384), and the Adaptive Masking Network (AMN) uses 4 Mamba layers.

Self-supervised pretraining spanned 2,400 epochs across three curriculum stages. Stage 1 (Foundational Pretraining) used public datasets for 600 epochs to learn general representations. Stage 2 (Robustness Pretraining) introduced synthetic clinical occlusions and ran for 1,000 epochs. Stage 3 (Domain-Specific Pretraining) involved 800 epochs on PICU data to align with clinical distribution.

Architectural and training hyperparameters were selected via systematic grid search and convergence analysis on a validation subset of the PICU datasets. We optimized the masking ratio, architecture depth, and stage durations to ensure stable learning and computational efficiency. Early stopping criteria were also considered to prevent overfitting during extensive training phases.

Masking ratios from 50\% to 95\% (in 5\% steps) were tested. A 75\% ratio was optimal: lower values (50–70\%) led to trivial reconstructions (MSE < 0.01), while higher values (80–95\%) caused instability in 30\% of training runs. A 75\% ratio also ensured visibility of at least one facial region in 90\% of frames and matched average occlusion severity in PICU videos (see Figure~\ref{fig:amn_temporal}).

For architecture depth, using more than 12 encoder layers yielded marginal gains (MAE improvement < 0.2 bpm) at a 25\% memory cost increase. Thus, we selected 12 encoder, 8 decoder, and 4 AMN layers. Increasing AMN depth by more than 4 layers led to 40\% longer training without significant masking accuracy improvement (only a 3\% gain in importance variance).

A dual-optimizer strategy was used. The student was trained using AdamW ($\beta_1 = 0.9$, $\beta_2 = 0.999$) with an initial learning rate of $1 \times 10^{-4}$ and cosine annealing, following a 40-epoch warm-up. AMN used a separate AdamW optimizer with a constant learning rate of $1 \times 10^{-5}$. Weight decay of 0.05 was applied to all trainable parameters except biases and normalization terms. Gradient clipping was enforced with a maximum $L_2$ norm of 1.0 for both optimizers to stabilize adversarial learning.

\subsubsection{Supervised Fine-tuning}

Following self-supervised pre-training, we perform supervised fine-tuning on annotated PICU data to enable accurate absolute heart rate estimation. From the full cohort of 500 patients, we selected a labeled subset of 200 patients. This subset was divided into 160 patients for training and 40 for testing. To enhance generalization, we conducted 5-fold patient-wise cross-validation across the 200 annotated patients. Each fold used 160 patients for training and 40 for testing, ensuring strict patient-level separation. All reported results reflect the average across folds, with 95\% confidence intervals.

To avoid data leakage, we enforced rigorous patient-level data partitioning. The 40 test patients were entirely excluded from all phases of training, including self-supervised pretraining. Specifically, during Stage 3 pretraining, we limited the PICU subset to 460 patients by excluding the 40 test patients and the 200 labeled patients reserved for evaluation. We also applied patient-independent clip splitting, ensuring that all 30-second clips from a single patient were assigned exclusively to either the training or test set. This protocol prevents the model from overfitting to patient-specific patterns and ensures generalizability across unseen subjects.

During supervised fine-tuning, the AMN is deactivated and bypassed. All spatial tokens are forwarded to the encoder without masking. Only the student encoder and the rPPG prediction head are updated during this stage. The loss function jointly aligns predicted and reference PPG signals and penalizes heart rate error:
\begin{equation}
    \mathcal{L}_{\text{supervised}} = 1 - \rho(\hat{\mathbf{y}}, \mathbf{y}_{\text{GT}}) + \lambda_{\text{HR}} \cdot |\text{HR}(\hat{\mathbf{y}}) - \text{HR}(\mathbf{y}_{\text{GT}})|
\end{equation}

where $\rho(\cdot, \cdot)$ denotes the Pearson correlation between the predicted signal $\hat{\mathbf{y}}$ and ground truth $\mathbf{y}{\text{GT}}$, and $\text{HR}(\cdot)$ computes heart rate using FFT peak detection in the 0.7–3.0 Hz frequency range. The loss weighting factor $\lambda_{\text{HR}}$ is set to 0.5.

Fine-tuning was performed using AdamW with a learning rate of $5 \times 10^{-5}$, set ten times lower than that used during pretraining. A cosine annealing schedule was applied across 100 epochs, with early stopping based on validation MAE. This phase refines the learned representations for precise clinical estimation while maintaining the robustness acquired through self-supervised learning.

\subsection{Evaluation metrics}

We report three quantitative metrics to assess heart rate estimation performance: mean absolute error (MAE), root mean squared error (RMSE), and Pearson correlation coefficient (R). All metrics are computed on a per-clip basis and averaged across the test set.

\begin{enumerate}
    \item Mean absolute error MAE: It is calculated by taking the average absolute difference between the predicted values $HR_{predict}$ and the ground truth values $HR_{reference}$:
        \begin{equation}MAE =\frac{1}{n} \sum_{i=1}^{n} | HR_{predict}^{i} -  HR_{reference}^{i} |.\label{MAE}\end{equation}
    
    \item Root mean squared error RMSE: It measures the average magnitude difference between the predicted value of heart rate and the actual one 
        \begin{equation}RMSE =\sqrt{\frac{1}{n} \sum_{i=1}^{n} (HR_{predict}^{i} -  HR_{reference}^{i})^2}.\label{RMSE}\end{equation}
        
    \item The Pearson correlation coefficient R : It measures the linear correlation between the two signals $rPPG_{predict}$ and $PPG_{reference}$: 
        \begin{equation}R =\frac{{\text{{cov}}(rPPG_{predict}, PPG_{reference})}}{{\text{{std}}(rPPG_{predict}) \cdot \text{{std}}(PPG_{reference})}}
    .\label{R}\end{equation}

\end{enumerate}

\section{Results and Discussion}

\begin{figure}[htbp]
    \centering
    \captionsetup[subfigure]{font=scriptsize, labelfont=scriptsize}
    
    \begin{subfigure}[b]{0.35\textwidth}
        \centering
        \includegraphics[width=\textwidth]{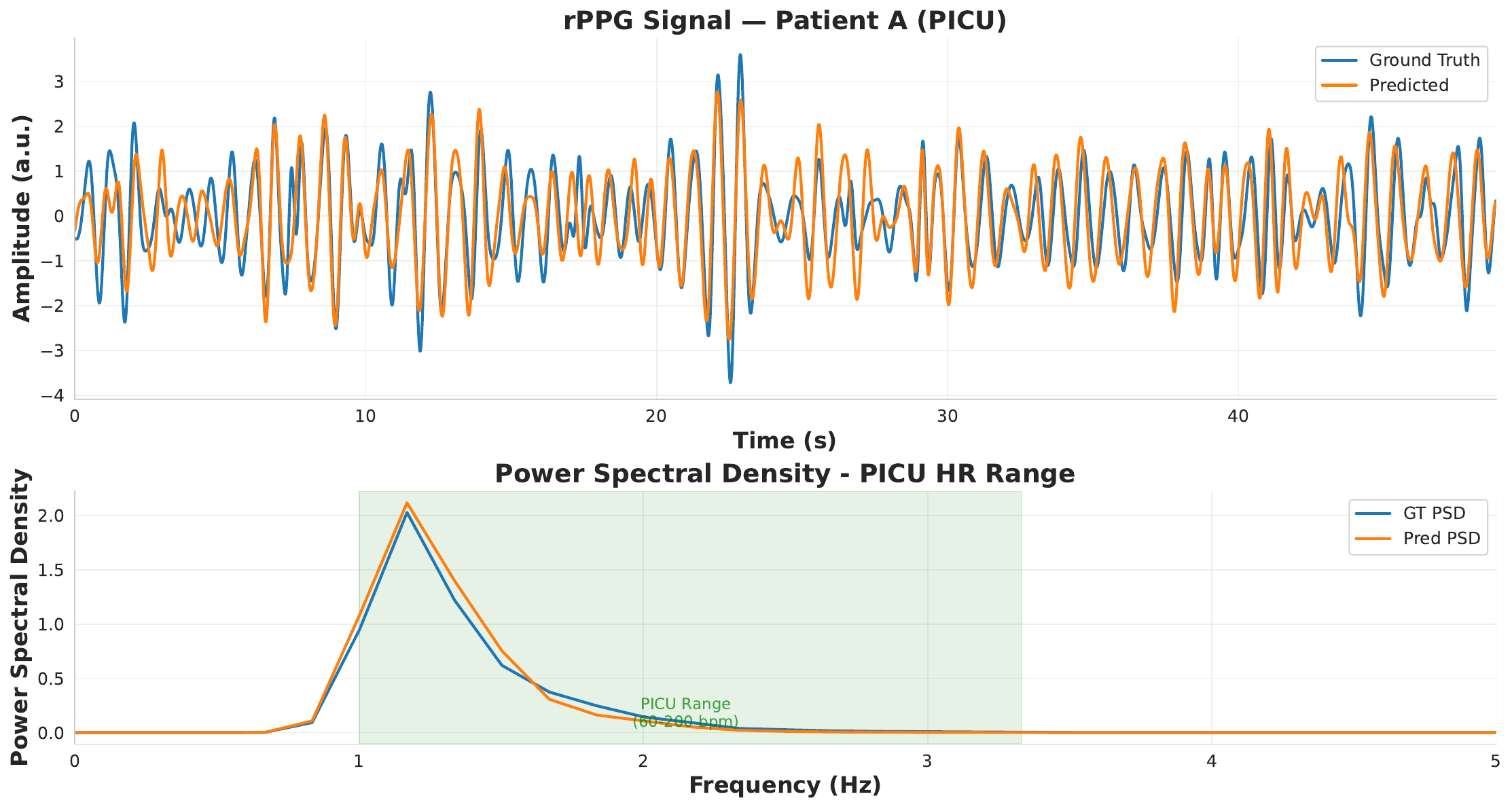}
        \caption{Patient A}
        \label{fig:qual_A}
    \end{subfigure}
    \hspace{3mm} 
    \begin{subfigure}[b]{0.35\textwidth}
        \centering
        \includegraphics[width=\textwidth]{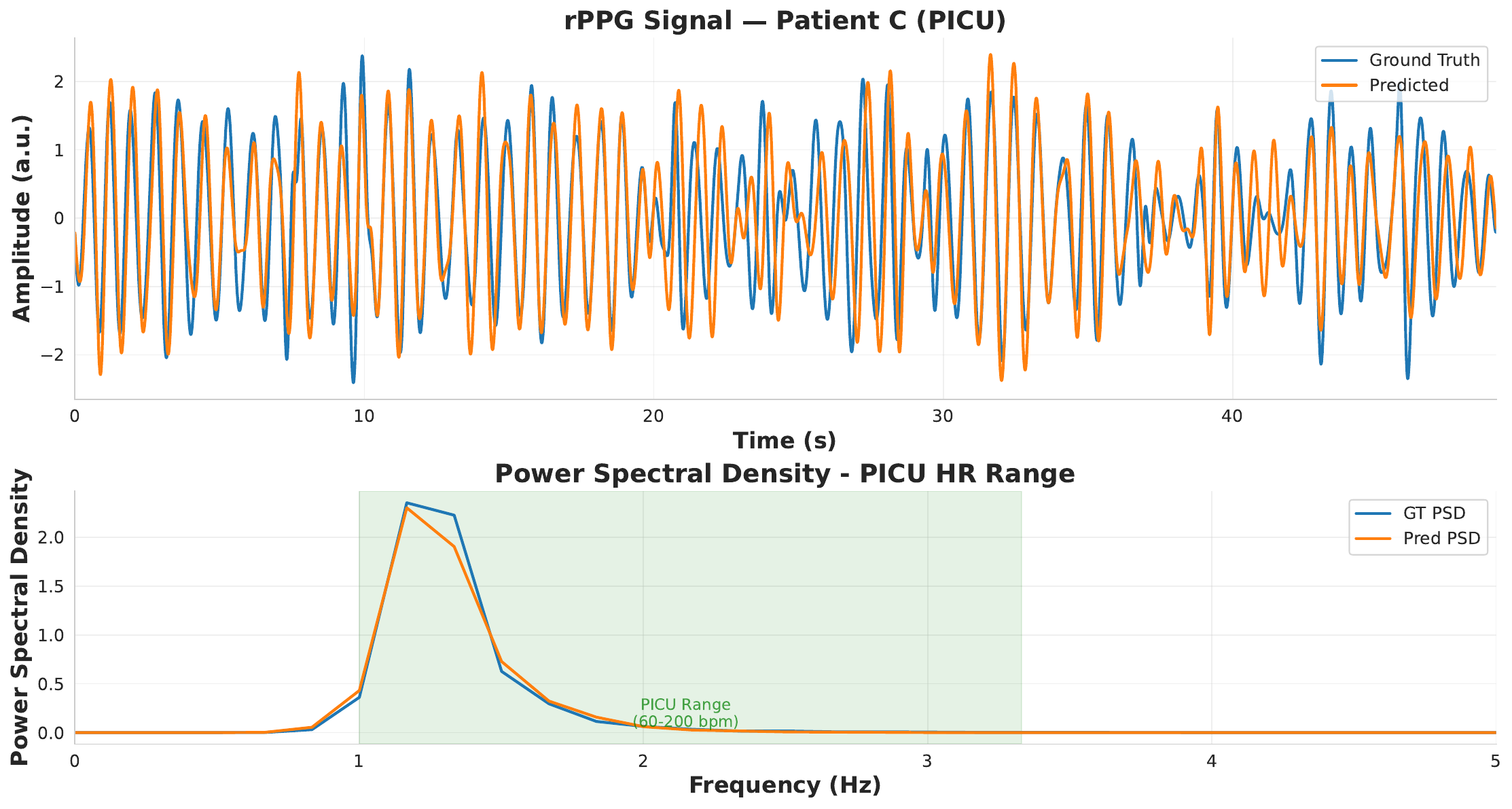}
        \caption{Patient C}
        \label{fig:qual_C}
    \end{subfigure}

    \vspace{1mm} 

    \begin{subfigure}[b]{0.35\textwidth}
        \centering
        \includegraphics[width=\textwidth]{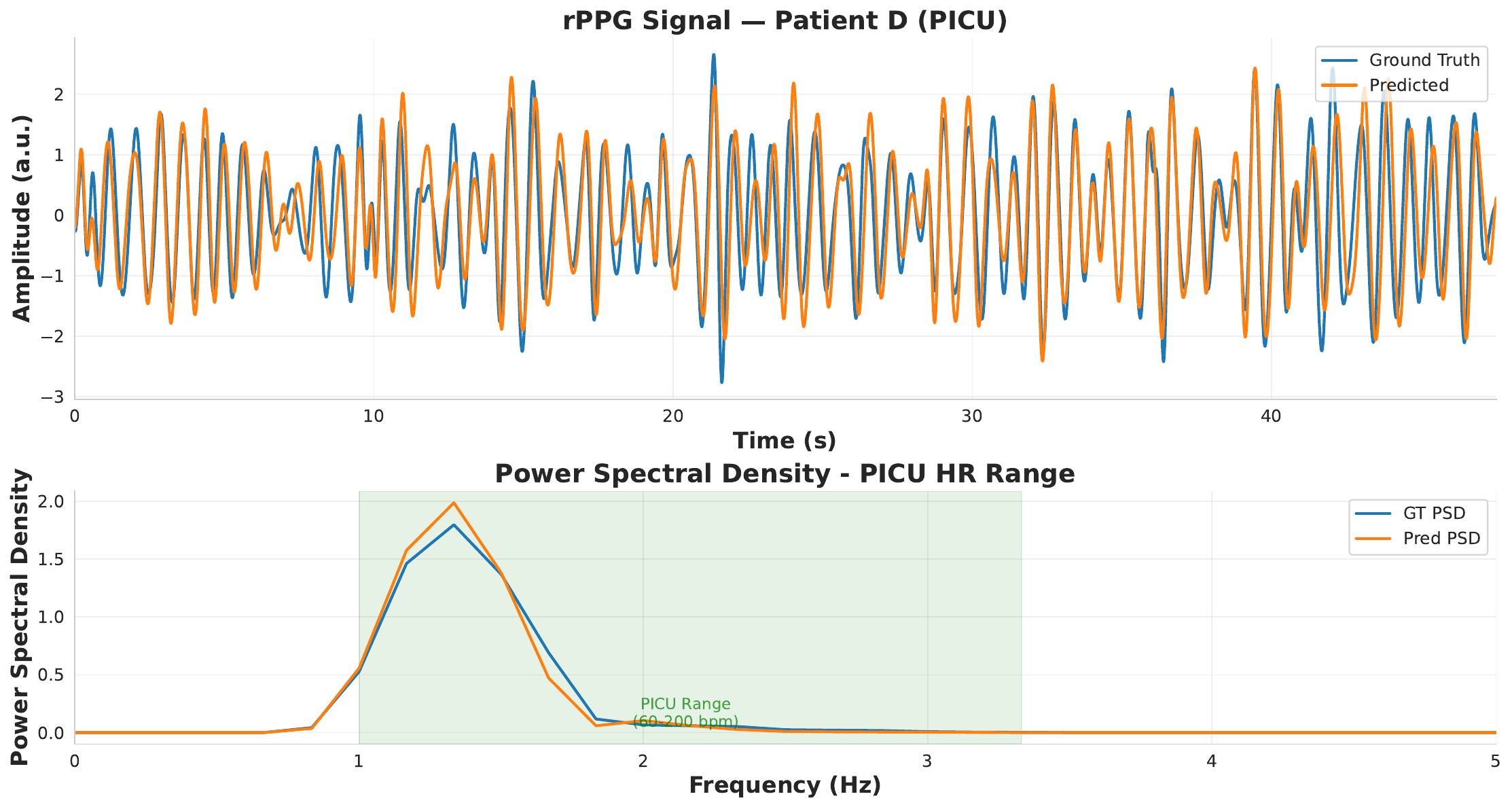}
        \caption{Patient D}
        \label{fig:qual_D}
    \end{subfigure}
    \hspace{3mm} 
    \begin{subfigure}[b]{0.35\textwidth}
        \centering
        \includegraphics[width=\textwidth]{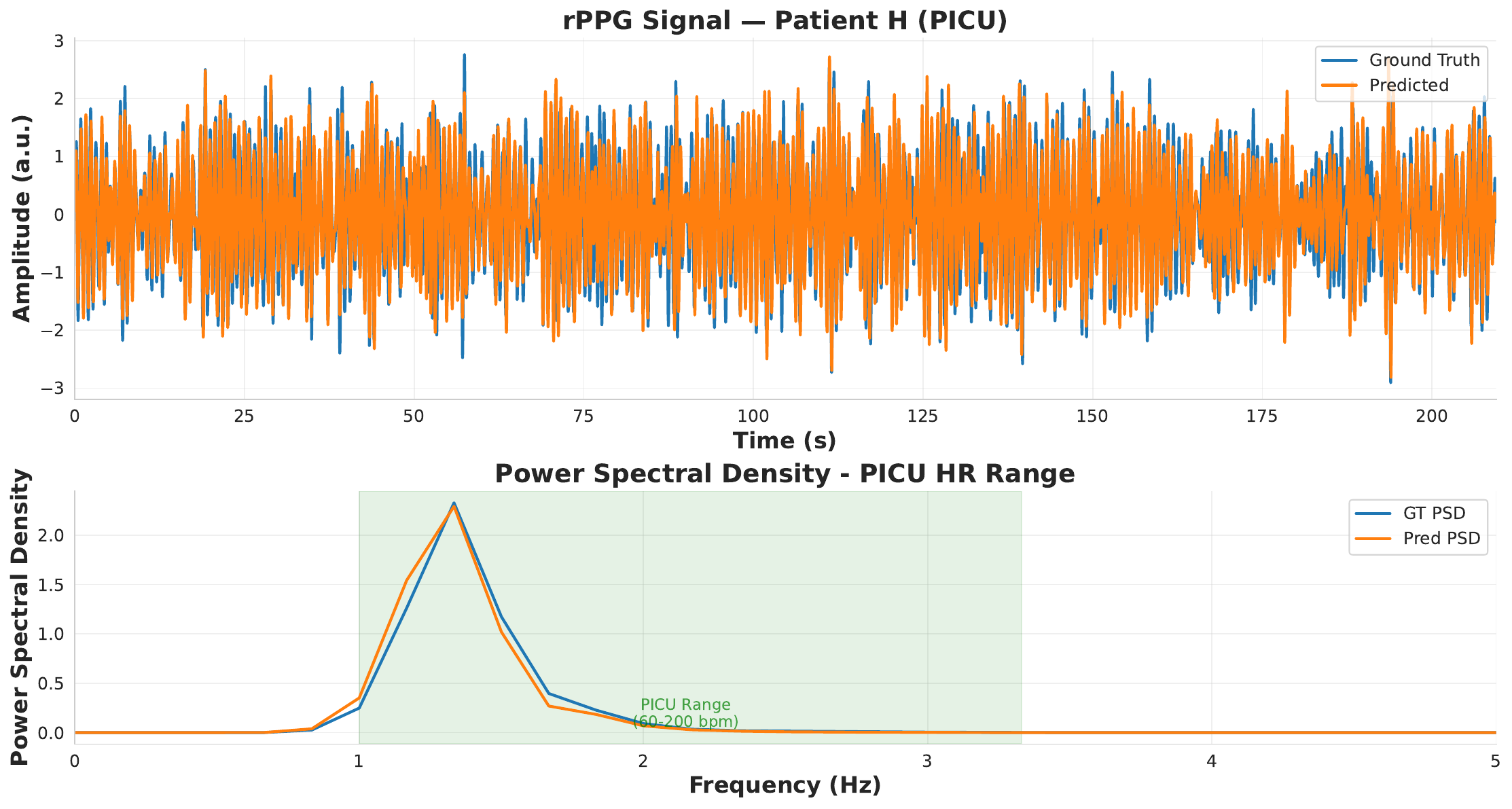}
        \caption{Patient H}
        \label{fig:qual_H}
    \end{subfigure}

    \caption{\small Qualitative results for select patients. Plots show reconstructed rPPG vs ground truth in time and frequency domains.}
    \label{fig:qualitative_signals_grid}
\end{figure}

\subsection{Comparison With State-of-the-Art Methods on the PICU Dataset}

%
   
  

\begin{table}[htbp]
  \centering
  \footnotesize 
  \renewcommand{\arraystretch}{1.1} 
  \setlength{\tabcolsep}{4pt} 
  
  \begin{minipage}[t]{0.48\textwidth}
    \centering
    \caption{Performance Comparison on PICU Test Set.}
    \label{tab:performance_picu_comparison}
    \begin{tabular}{lcccc}
      \toprule
      \textbf{Method} & \textbf{MAE} & \textbf{RMSE} & \textbf{MPE} & \textbf{R} \\
      \midrule
      CHROM~\cite{bib9} & 15.8 & 19.2 & 18.3 & 0.42 \\
      POS~\cite{bib10}   & 14.2 & 17.8 & 16.7 & 0.48 \\
      \midrule
      PhysNet~\cite{bib17}       & 8.7 & 11.3 & 10.2 & 0.71 \\
      EfficientPhys~\cite{bib14} & 7.9 & 10.4 & 9.3 & 0.74 \\
      PhysMamba~\cite{bib7}\dag  & 7.2 & 9.8 & 8.5 & 0.77 \\
      MTTS-CAN~\cite{bib15}      & 6.5 & 8.9 & 7.6 & 0.81 \\
      PhysFormer~\cite{bib28}    & 5.8 & 8.2 & 6.8 & 0.84 \\
      \midrule
      Contrast-Phys~\cite{Sun2022ContrastPhys} & 11.3 & 14.2 & 13.1 & 0.58 \\
      rPPG-MAE~\cite{Liu2023rPPGMAE}          & 9.6 & 12.4 & 11.2 & 0.66 \\
      \midrule
      Ours (Base)  & 10.3 & 17.6 & 14.2 & 0.67 \\
      \textbf{Ours (Final)} & \textbf{3.2} & \textbf{5.4} & \textbf{3.8} & \textbf{0.91} \\
      \bottomrule
      \multicolumn{5}{l}{\dag Teacher model used in framework} \\
    \end{tabular}
  \end{minipage}
  \hfill 
  \begin{minipage}[t]{0.48\textwidth}
    \centering
    \caption{Model Performance Across Patient Subgroups.}
    \label{tab:performance_demographics}
    \begin{tabular}{lcccc}
      \toprule
      \textbf{Subgroup} & \textbf{N} & \textbf{MAE} & \textbf{RMSE} & \textbf{R} \\
      \midrule
      \multicolumn{5}{l}{\textbf{Age Groups}} \\
      Neonates    & 35 & 3.8 & 5.9 & 0.82 \\
      Infants     & 77 & 3.4 & 5.5 & 0.87 \\
      Toddlers    & 27 & 3.1 & 5.2 & 0.91 \\
      Children    & 12 & 2.9 & 5.0 & 0.93 \\
      Adolescents & 8  & 1.7 & 3.8 & 0.95 \\
      \midrule
      \multicolumn{5}{l}{\textbf{Skin Tone (Fitzpatrick)}} \\
      Type I-II   & 14 & 2.8 & 4.9 & 0.92 \\
      Type III-IV & 18 & 3.2 & 5.4 & 0.91 \\
      Type V-VI   & 8  & 3.9 & 6.1 & 0.88 \\
      \midrule
      \multicolumn{5}{l}{\textbf{Clinical Condition}} \\
      Ventilated     & 24 & 3.5 & 5.7 & 0.87 \\
      Non-Ventilated & 16 & 2.7 & 4.9 & 0.93 \\
      \bottomrule
    \end{tabular}
  \end{minipage}
\end{table}

As shown in Table~\ref{tab:performance_picu_comparison}, the fine-tuned model achieves a MAE of 3.2~bpm and RMSE of 5.4~bpm on the PICU test set, surpassing both traditional and deep learning approaches. Compared to PhysFormer (MAE: 5.8~bpm), the proposed method reduces the error by 44.8\% and achieves a 50.8\% improvement over MTTS-CAN (MAE: 6.5~bpm). The Pearson correlation increases from 0.84 to 0.91, indicating closer alignment between the predicted and reference waveforms.

Without supervised fine-tuning, the self-supervised model already attains a MAE of 10.3~bpm, outperforming Contrast-Phys (11.3~bpm) and approaching the performance of early supervised architectures. Fine-tuning on 160 labeled PICU patients further reduces the error from 10.3~bpm to 3.2~bpm, representing a relative improvement of 68.9\%. This demonstrates that the pretraining stage provides a strong initialization that captures physiologically meaningful temporal features, while supervised adaptation refines these representations for clinical precision.

This combination yields robust performance in the presence of domain-specific challenges such as motion, occlusion, and poor illumination. Fine-tuning on 160 annotated patients provides a substantial reduction in error and confirms that the pretraining strategy learns meaningful spatiotemporal dynamics from large-scale unlabeled PICU data.

\subsection{Waveform Analysis of rPPG Signal Reconstruction}

Figure~\ref{fig:qualitative_signals_grid} shows qualitative results from patients in the PICU cohort, comparing predicted rPPG signals and corresponding power spectral density (PSD) curves to ground truth references. In the time domain, the model preserves the pulsatile morphology and beat-to-beat variability. In the frequency domain, the PSD curves exhibit sharp, localized peaks aligned with the ground truth heart rate, indicating robustness to motion, occlusion, and subject variability.

These observations align with the quantitative results in Table~\ref{tab:performance_picu_comparison}, and further support the model's ability to recover physiologically meaningful signals without relying on fixed regions of interest or landmark-based supervision. The framework maintains performance even under occlusion, leveraging spatiotemporal dynamics learned during training.

The final example in Figure~\ref{fig:qualitative_signals_grid} corresponds to recordings longer than 210 seconds. The reconstructed signals preserve regular pulsatile structure across the full duration, and the associated PSD curves remain confined within the expected pediatric heart rate range (60–200 bpm, or 1.0–3.3 Hz). This illustrates the model's capacity to support continuous monitoring where long-term signal stability is critical.

\begin{figure}[htbp]
    \centering
    \captionsetup[subfigure]{font=scriptsize, labelfont=scriptsize}
    
    \begin{subfigure}[b]{0.35\textwidth}
        \centering
        \includegraphics[width=\textwidth]{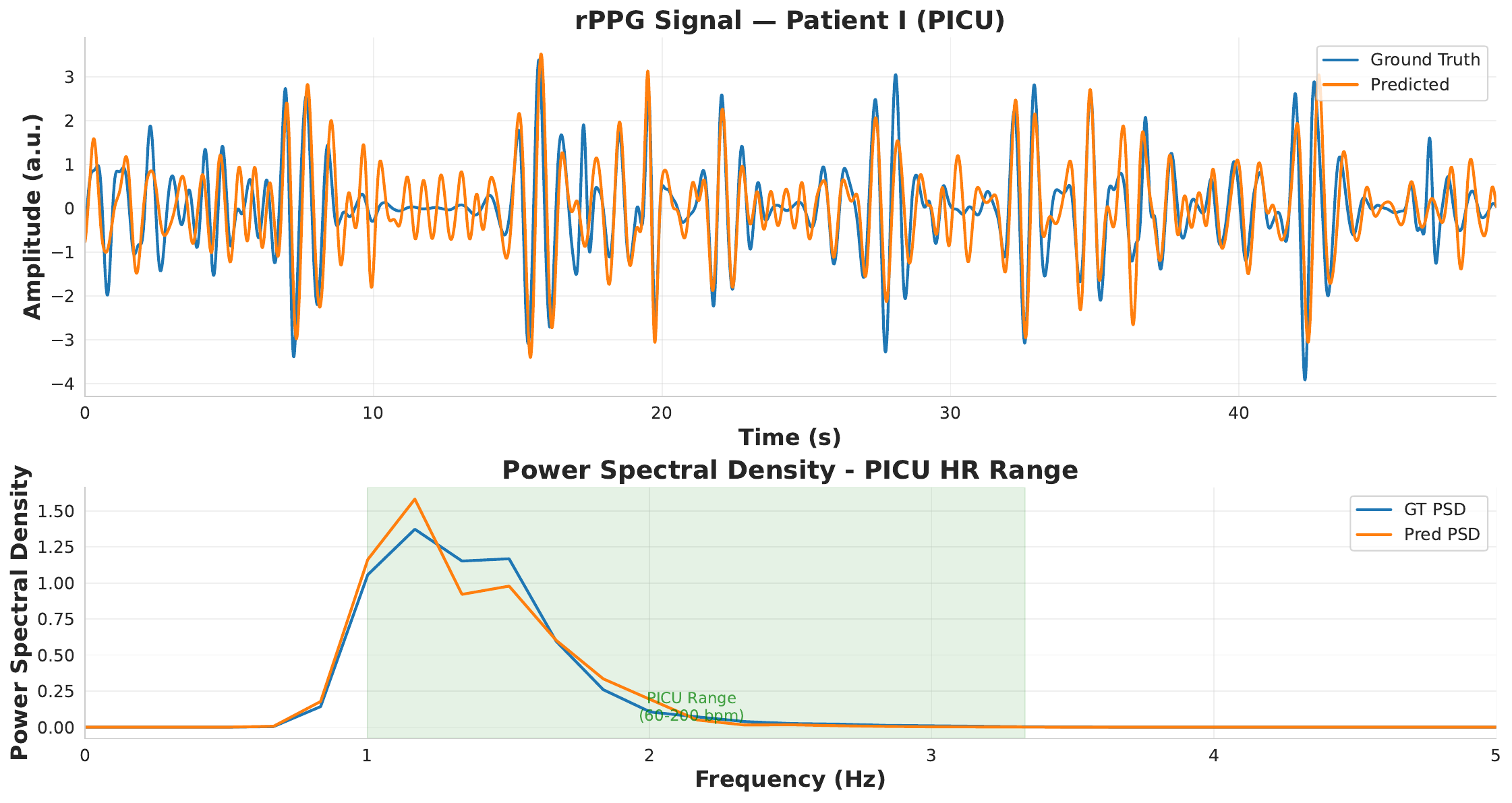}
        \caption{Patient I}
        \label{fig:patient_i}
    \end{subfigure}
    \hspace{8mm} 
    \begin{subfigure}[b]{0.35\textwidth}
        \centering
        \includegraphics[width=\textwidth]{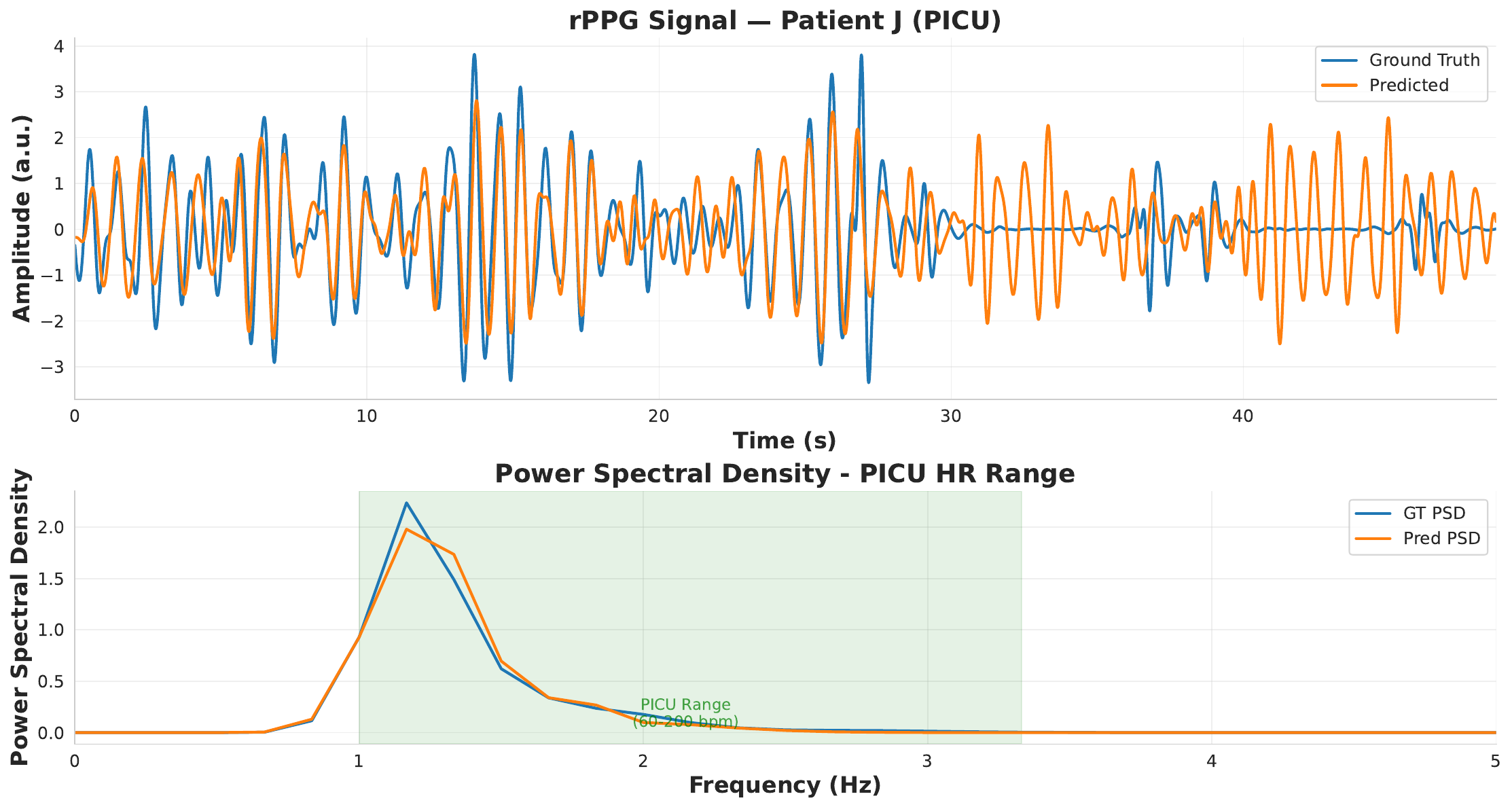}
        \caption{Patient J}
        \label{fig:patient_j}
    \end{subfigure}
    
    \caption{\small Resilience to sensor artifacts. The model maintains stable rPPG waveforms even during periods of ground-truth PPG signal degradation (flat-lines).}
    \label{fig:resilience_rppg}
\end{figure}

Figure~\ref{fig:resilience_rppg} highlights cases where the model compensates for reference signal artifacts. In many PICU recordings, the contact-based oximeter shows flat segments or discontinuities due to disconnection or motion. Despite these interruptions, the predicted rPPG signals remain regular, and physiologically plausible.

For example, in Patient I (35–40~s) and J (20–25~s), the reference PPG signal exhibits abrupt degradation, while the model output maintains continuous pulse waveforms. This highlights a clinical advantage: unlike contact sensors that are prone to physical failure, the proposed method provides stable monitoring even under challenging acquisition conditions.

In the frequency domain, the reconstructed PSD curves retain sharp peaks within the expected heart rate range, confirming that the model preserves rhythmic content even when the reference signal is unreliable or missing.

\subsection{Performance Across Different Patient Subgroups}

%

As summarized in Table~\ref{tab:performance_demographics}, the model achieves consistent performance across all age subgroups. The MAE decreases progressively with age, from 3.8~bpm in neonates to 1.7~bpm in adolescents. This trend indicates stable generalization across physiological variability and developmental motion patterns.

When stratified by skin tone, the model shows slightly higher error for Fitzpatrick Types~V–VI (3.9~bpm) compared to Types~I–II (2.8~bpm), a difference of 1.1~bpm. This aligns with prior findings on reduced signal-to-noise ratio in rPPG estimation due to melanin absorption~\cite{Dasari2021Biases, Talukdar2023Evaluation, Nowara2020Meta}.

Performance also differs by ventilation status. Mechanically ventilated patients present an MAE of 3.5~bpm, while non-ventilated patients achieve 2.7~bpm. The presence of medical devices such as tubing or adhesive patches may explain this difference by introducing occlusions or artifacts. Despite these subgroup differences, Pearson correlation remains high (0.87 to 0.95), confirming robust temporal alignment with ground truth heart rate.

\subsection{Robustness to Clinical Occlusions}

\begin{figure}[htbp]
    \centering
    \includegraphics[width=0.45\textwidth]{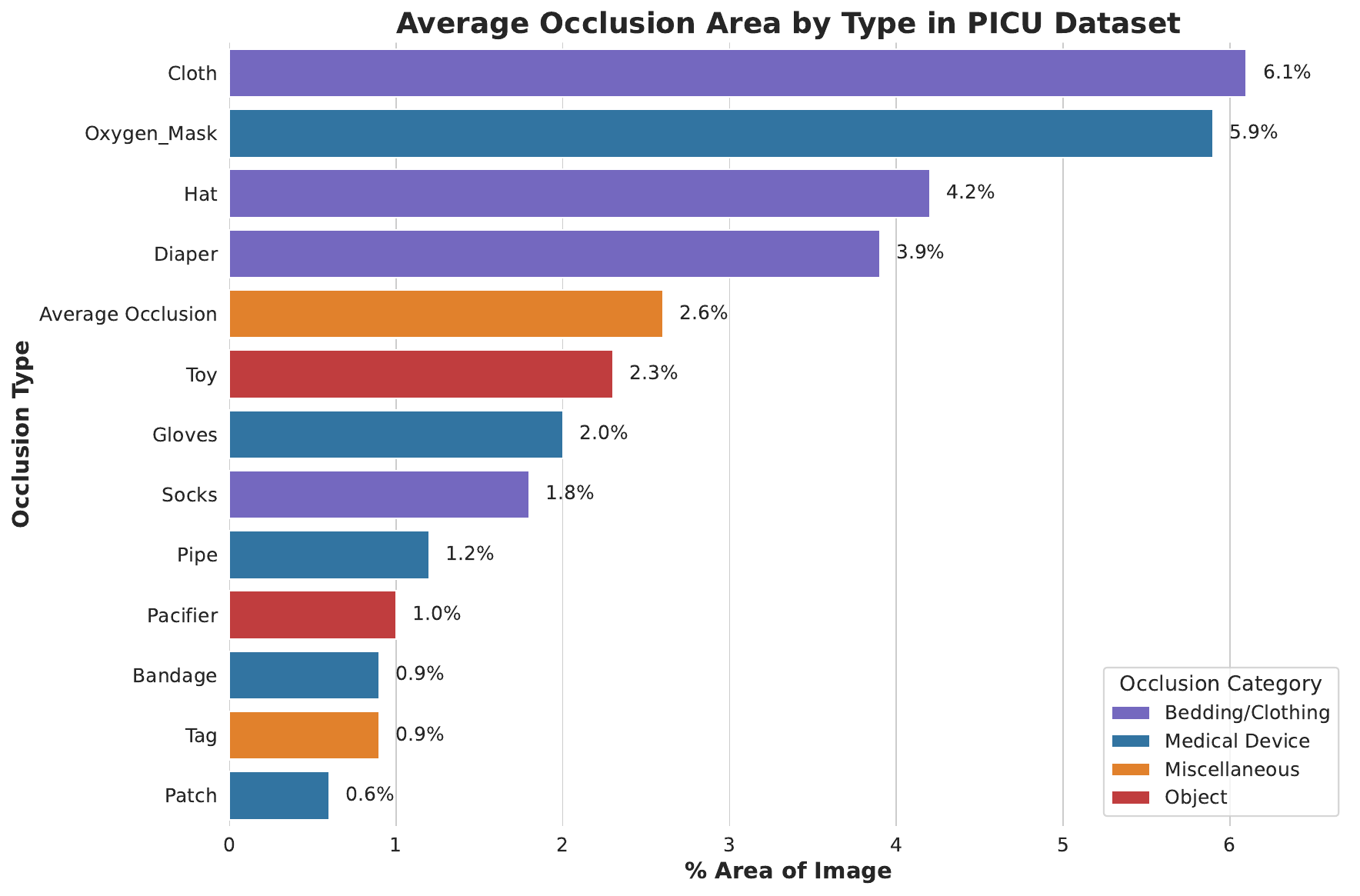}
    \caption{\footnotesize Average occlusion area by type. Medical devices and bedding represent the most significant sources of signal obstruction.}
    \label{fig:occlusion_types}
\end{figure}

We assessed model performance under common occlusion scenarios encountered in the PICU environment. As shown in Figure~\ref{fig:occlusion_types}, the most frequent sources of obstruction include medical devices and bedding. Oxygen masks and cloth coverings are particularly prominent, occluding an average of 6.1\% and 5.9\% of the image area, respectively. These occlusions often affect key facial regions required for accurate rPPG estimation.

Despite these challenges, the proposed model maintains stable performance across all occlusion categories and consistently outperforms baseline methods (Table~\ref{tab:performance_occlusion}). In cases of severe occlusion (greater than 70\% of the facial area), our method achieves a mean absolute error (MAE) of 7.2~bpm, compared to 13.3~bpm with PhysFormer and 16.8~bpm with MTTS-CAN. These values correspond to relative error reductions of 45.9\% and 57.1\%, respectively. Under moderate occlusion (25 to 70\%), the MAE reaches 3.8~bpm, while PhysFormer and MTTS-CAN yield 8.7~bpm and 10.2~bpm, respectively, resulting in reductions of 56.3\% and 62.7\%. Similar trends are observed in light and minimal occlusion conditions. These findings underscore the ability of our adaptive masking and distillation strategy to maintain signal integrity in the presence of visual obstructions.

\begin{table}[htbp]
  \centering
  \footnotesize
  \renewcommand{\arraystretch}{1.1} 
  \setlength{\tabcolsep}{3.5pt}     

  \begin{minipage}[t]{0.48\textwidth}
    \vspace{0pt} 
    \centering
    \caption{Performance Under Different Occlusion Levels.}
    \label{tab:performance_occlusion}
    \begin{tabular}{lccc}
      \toprule
      \textbf{Level} & \textbf{Ours} & \textbf{PhysF.} & \textbf{MTTS} \\
      \midrule
      None ($<$10\%)    & \textbf{2.3} & 4.8 & 5.9 \\
      Light (10--25\%)  & \textbf{2.9} & 6.1 & 7.4 \\
      Mod. (25--70\%)   & \textbf{3.8} & 8.7 & 10.2 \\
      Sev. ($>$70\%)    & \textbf{7.2} & 13.3 & 16.8 \\
      \bottomrule
    \end{tabular}
  \end{minipage}
  \hfill
  \begin{minipage}[t]{0.48\textwidth}
    \vspace{0pt} 
    \centering
    \caption{UBFC-Occluded Results (20s windows).}
    \label{tab:ubfc_occ_simple}
    \begin{tabular}{lccc}
      \toprule
      \textbf{Stage} & \textbf{MAE}$\downarrow$ & \textbf{RMSE}$\downarrow$ & \boldmath$R$\unboldmath$\uparrow$ \\
      \midrule
      Stage 1 only   & 5.4 & 13.0 & 0.76 \\
      Stage 1+2      & 3.1 & 4.7  & 0.89 \\
      \textbf{Full Pipeline} & \textbf{2.8} & \textbf{4.1} & \textbf{0.93} \\
      \bottomrule
    \end{tabular}
  \end{minipage}
\end{table}

\subsection{Cross-dataset Evaluation on UBFC with Synthetic Occlusions}

\begin{figure}[htbp]
    \centering
    \includegraphics[width=0.35\textwidth]{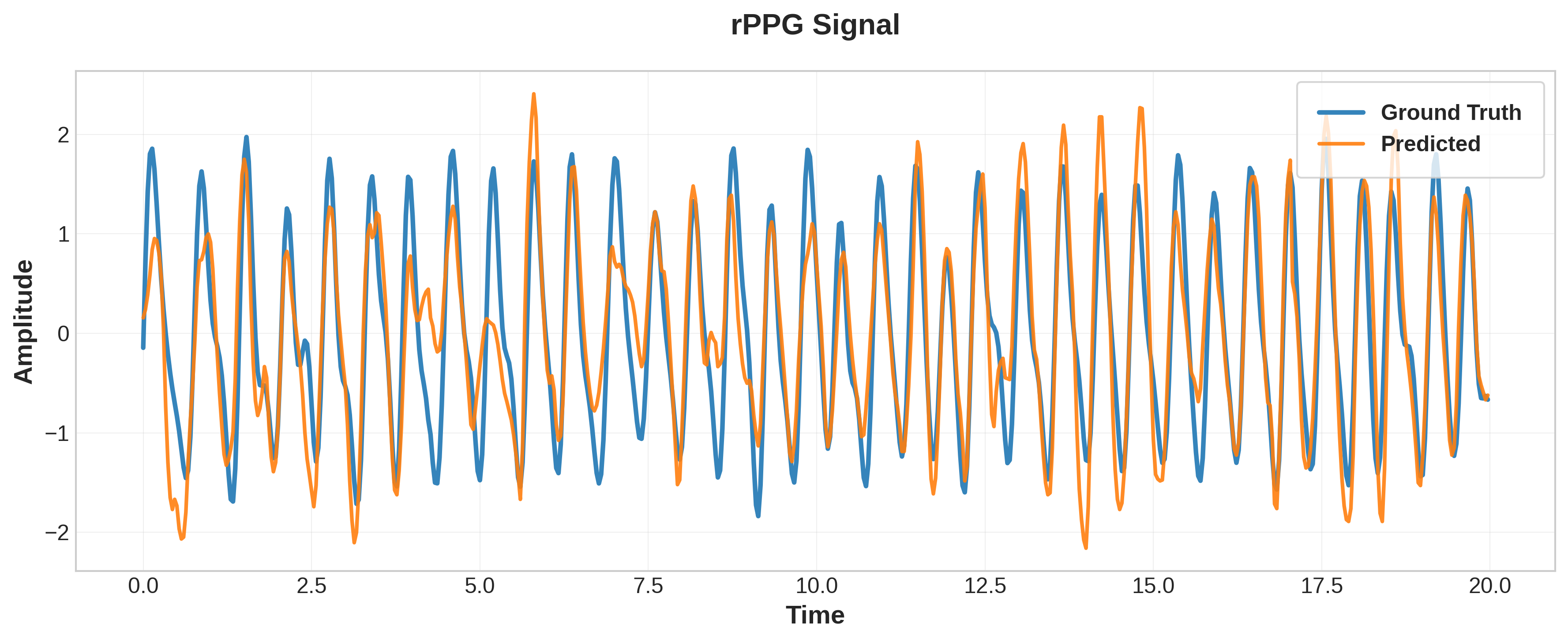}
    \hspace{5mm} 
    \includegraphics[width=0.35\textwidth]{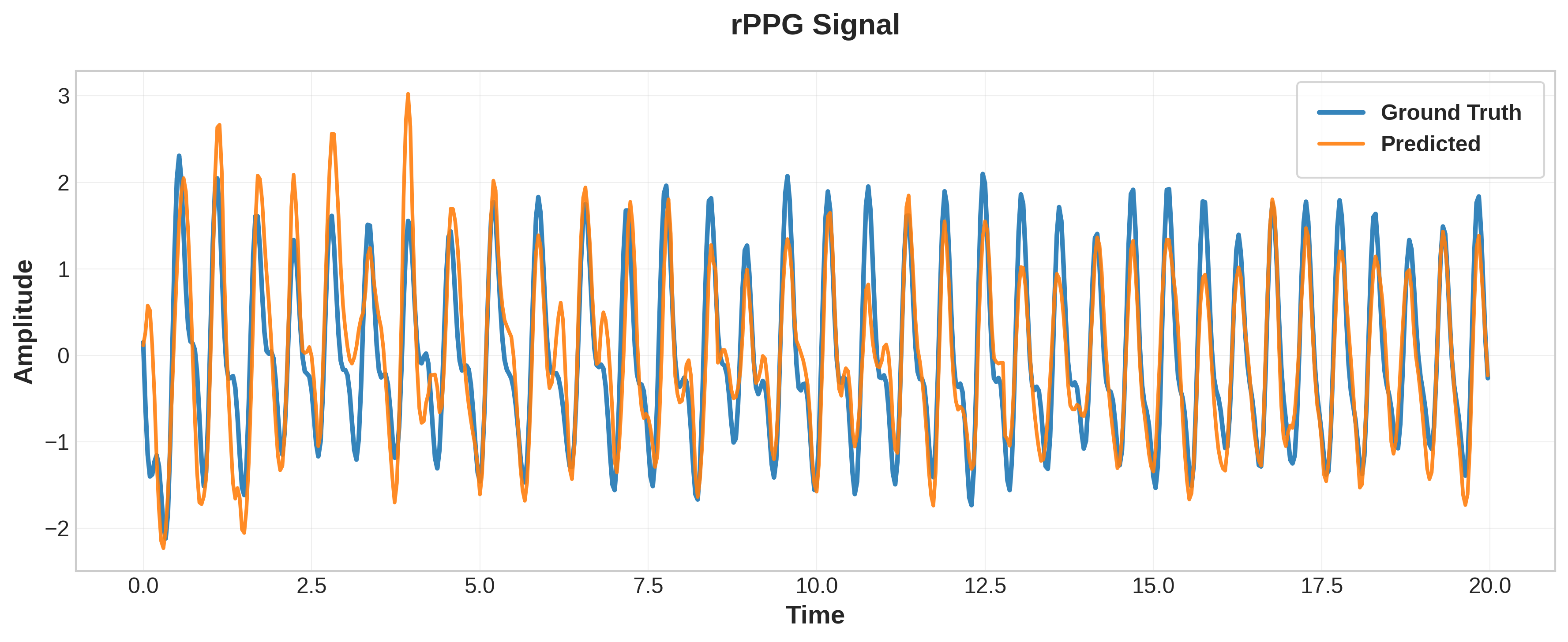}

    \caption{\footnotesize Time-domain overlay of predicted and ground truth rPPG signals. This compact view shows the model's tracking accuracy on UBFC subjects under synthetic occlusion scenarios.}
    \label{fig:qualitative_rppg}
\end{figure}

We evaluated the staged pretraining strategy on the UBFC-rPPG dataset under controlled synthetic occlusions to examine the model’s ability to generalize across domains. As presented in Table~\ref{tab:ubfc_occ_simple}, the heart rate estimation error progressively decreases over the three stages (MAE: 5.4~$\rightarrow$~3.1~$\rightarrow$~2.8~bpm). The addition of Stage~2, which introduces occlusion augmentation, results in a significant reduction in error, suggesting that the training pipeline enhances robustness to obstructed facial regions and improves generalization to clinical-like conditions.

Figure~\ref{fig:qualitative_rppg} displays examples from four UBFC subjects with 40 to 60\% facial occlusion, including simulated masks, tubing, and overlays. Despite reduced visibility, the predicted rPPG signals remain temporally consistent and closely follow the ground truth. These qualitative observations reinforce the quantitative findings and demonstrate that curriculum-guided self-supervised learning improves model reliability under occlusion and domain shift.

\subsection{Qualitative Analysis and Model Interpretability}

\begin{figure}[htbp]
    \centering
    \captionsetup[subfigure]{font=scriptsize, labelfont=scriptsize}

    \begin{subfigure}[b]{0.28\textwidth}
        \includegraphics[width=\textwidth, keepaspectratio]{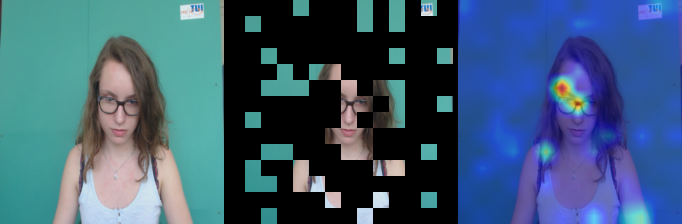}
    \end{subfigure}
    \hfill
    \begin{subfigure}[b]{0.28\textwidth}
        \includegraphics[width=\textwidth, keepaspectratio]{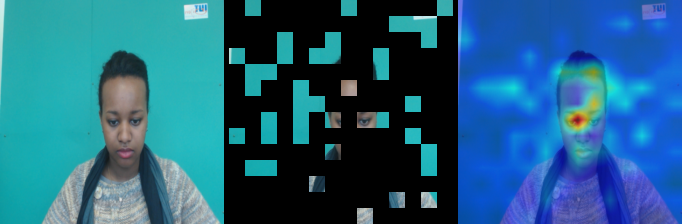}
    \end{subfigure}
    \hfill
    \begin{subfigure}[b]{0.28\textwidth}
        \includegraphics[width=\textwidth, keepaspectratio]{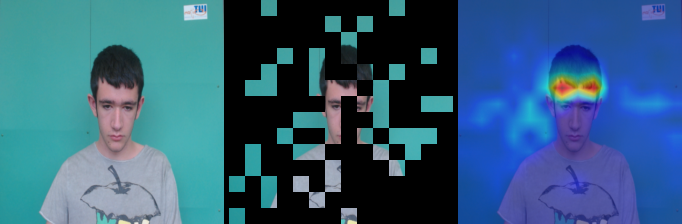}
    \end{subfigure}

    \vspace{1mm} 

    \begin{subfigure}[b]{0.28\textwidth}
        \includegraphics[width=\textwidth, keepaspectratio]{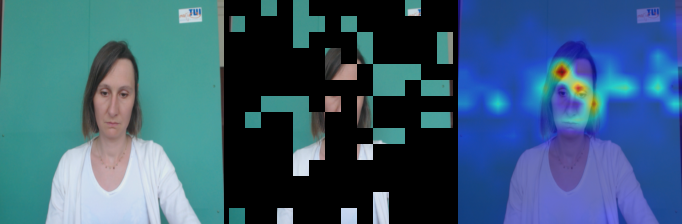}
    \end{subfigure}
    \hfill
    \begin{subfigure}[b]{0.28\textwidth}
        \includegraphics[width=\textwidth, keepaspectratio]{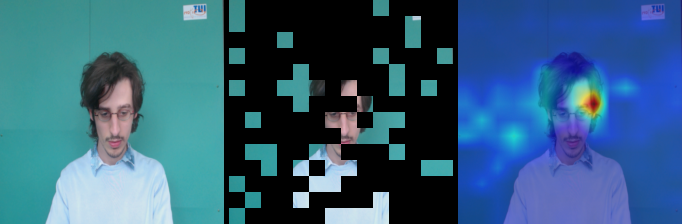}
    \end{subfigure}
    \hfill
    \begin{subfigure}[b]{0.28\textwidth}
        \includegraphics[width=\textwidth, keepaspectratio]{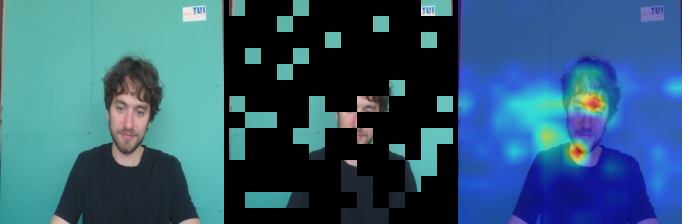}
    \end{subfigure}

    \caption{\footnotesize Grad-CAM spatial attention on six subjects. Triplets show original, 75\% masked, and heatmap outputs. Despite heavy masking, focus consistently aligns with vascular regions.}
    \label{fig:grad_cam_vis}
\end{figure}

Figure~\ref{fig:grad_cam_vis} presents Gradient-weighted Class Activation Mapping (Grad-CAM) visualizations from six UBFC subjects to examine the spatial attention learned by the student model. For each subject, three views are shown: the original RGB frame, the masked input with 75\% token dropout from the Adaptive Masking Network (AMN), and the Grad-CAM activation map extracted from the final rPPG decoder layer.

Despite strong masking during pretraining, the model consistently focuses on physiologically relevant facial regions, particularly the forehead, periorbital areas, and upper cheeks. These regions correspond to areas with dense superficial vasculature, such as the temporal and facial arteries, which are optimal for pulse extraction. This spatial selectivity emerges without explicit supervision or predefined regions of interest, indicating that the combination of reconstruction objectives and distillation effectively encodes physiological priors.

The masking strategy contributes significantly to this selective attention behavior. By intentionally suppressing informative tokens during pretraining, the AMN forces the model to rely on the remaining visible facial regions for pulse recovery. Grad-CAM analysis shows that the model adapts dynamically, shifting its focus toward alternative vascular regions when primary areas are occluded, while maintaining physiological consistency. This adaptive mechanism directly supports the quantitative findings in Table~\ref{tab:performance_occlusion}, where the model retains a MAE of 5.2~bpm under more than 50\% occlusion, compared to 14.3~bpm for PhysFormer.

The visualizations also reveal strong spatial selectivity, with minimal activation in background and masked regions. The model concentrates attention on exposed skin areas, independent of skin tone, facial structure, or the presence of accessories such as glasses. This suggests that attention is guided by physiological cues rather than superficial appearance.

These qualitative observations support the hypothesis that combining adaptive masking with physiological distillation enables the model to identify clinically relevant features without handcrafted priors. The model also maintains accurate pulse estimation under severe occlusions. This performance, achieved through self-supervised pretraining followed by limited supervised fine-tuning, confirms the ability of the framework to learn robust and physiologically grounded representations.

\begin{figure}[htbp]
\centering
\captionsetup[subfigure]{font=tiny, labelfont=tiny}

\begin{subfigure}[b]{0.09\textwidth}
  \includegraphics[width=\textwidth]{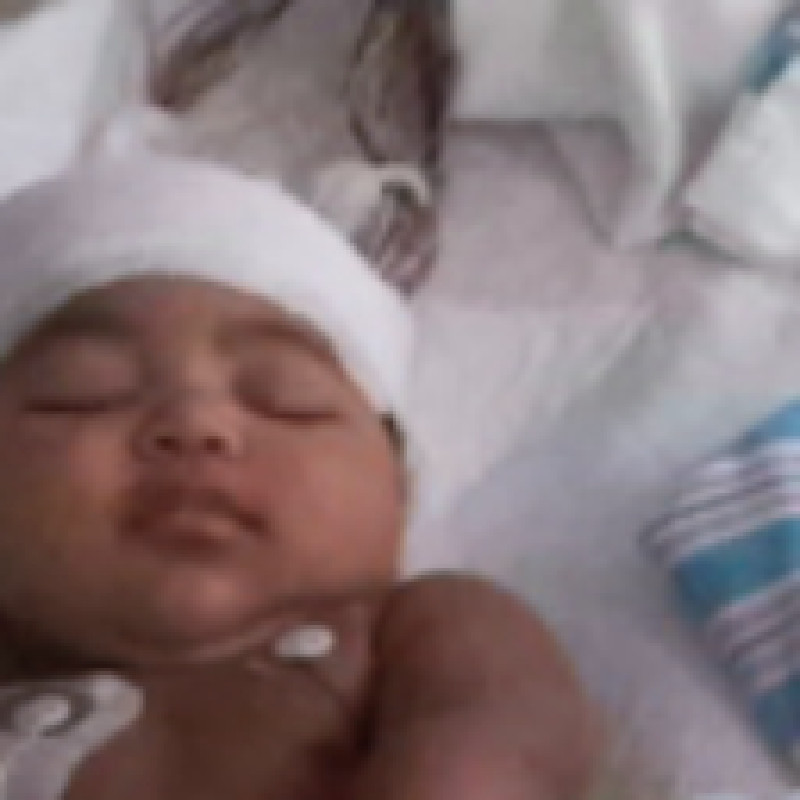}\\[-1pt]
  \includegraphics[width=\textwidth]{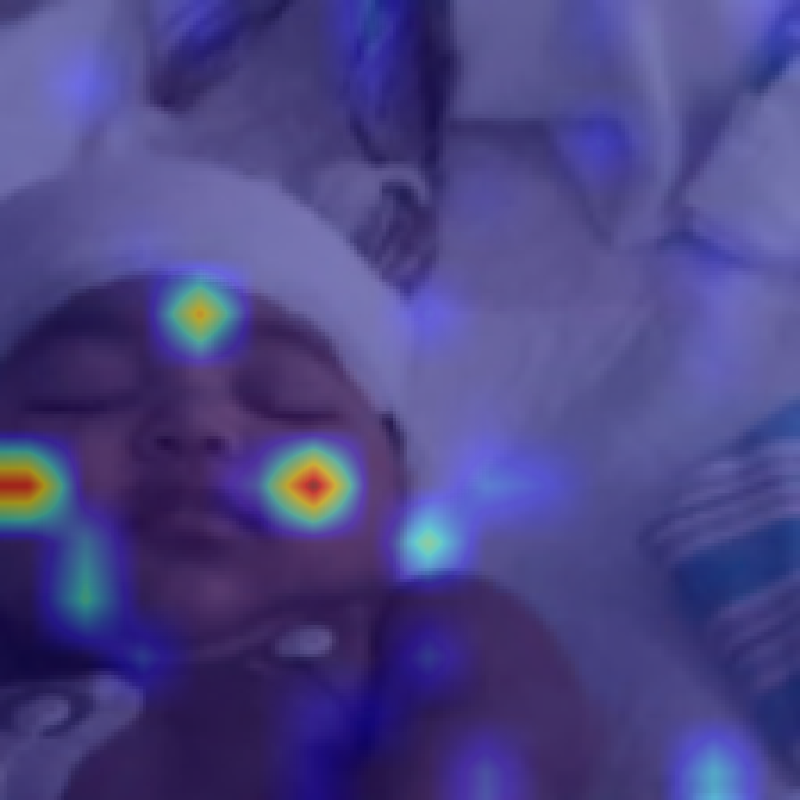}
  \caption*{P1}
\end{subfigure}\hfill
\begin{subfigure}[b]{0.09\textwidth}
  \includegraphics[width=\textwidth]{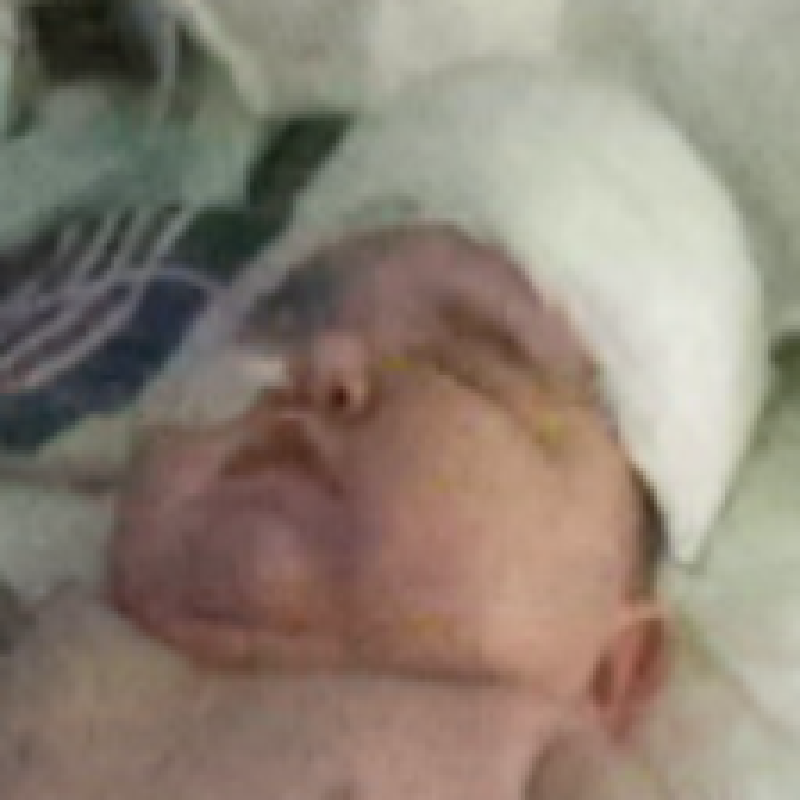}\\[-1pt]
  \includegraphics[width=\textwidth]{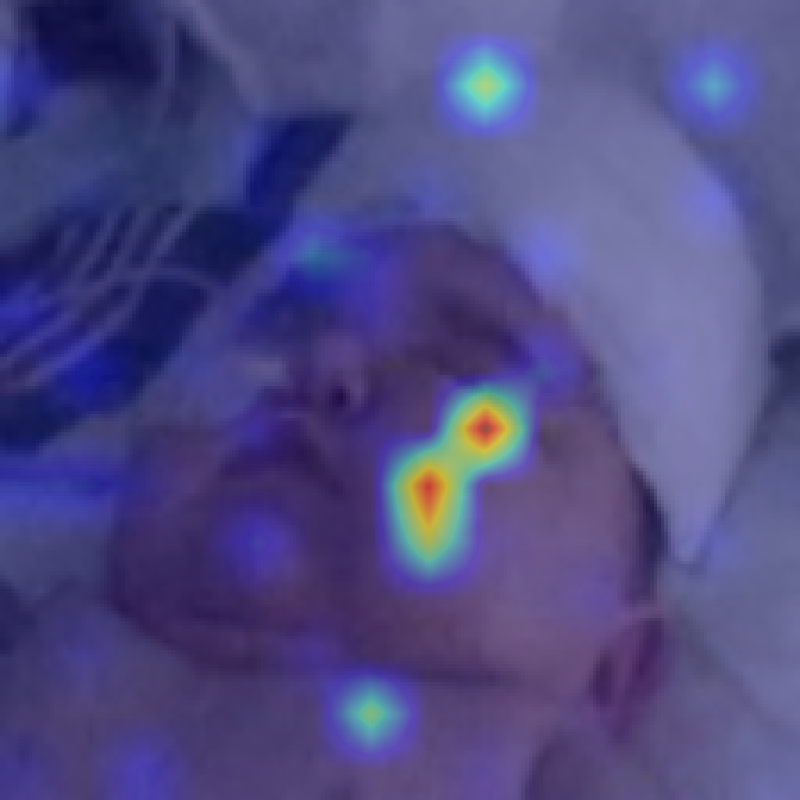}
  \caption*{P2}
\end{subfigure}\hfill
\begin{subfigure}[b]{0.09\textwidth}
  \includegraphics[width=\textwidth]{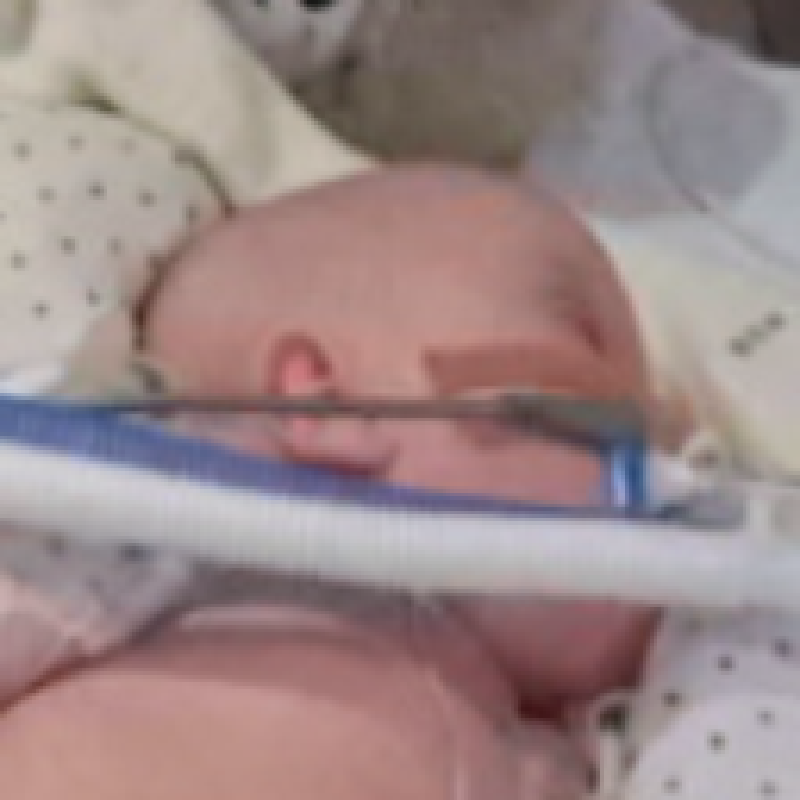}\\[-1pt]
  \includegraphics[width=\textwidth]{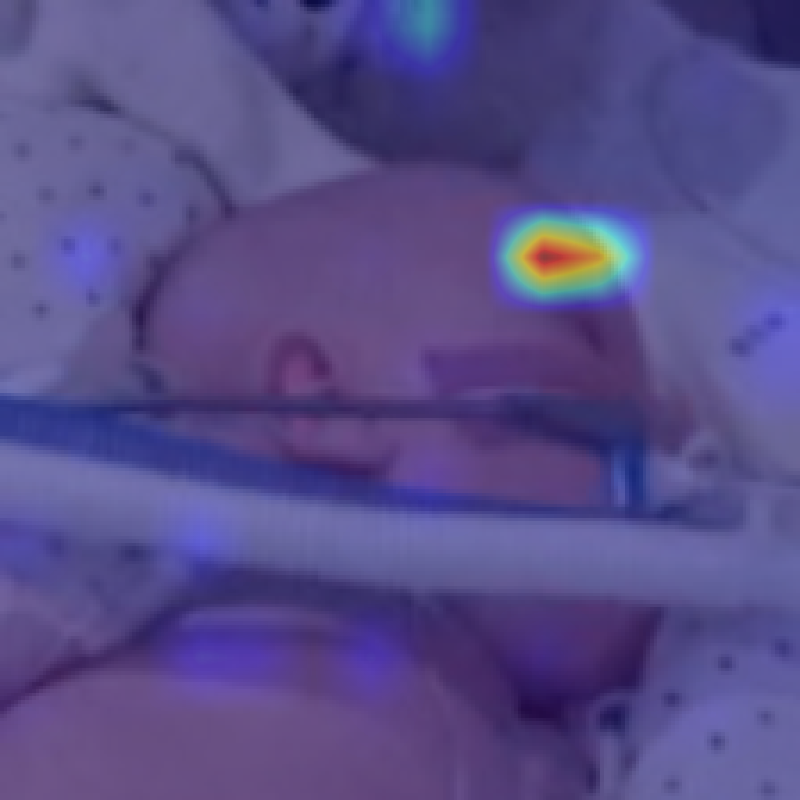}
  \caption*{P3}
\end{subfigure}\hfill
\begin{subfigure}[b]{0.09\textwidth}
  \includegraphics[width=\textwidth]{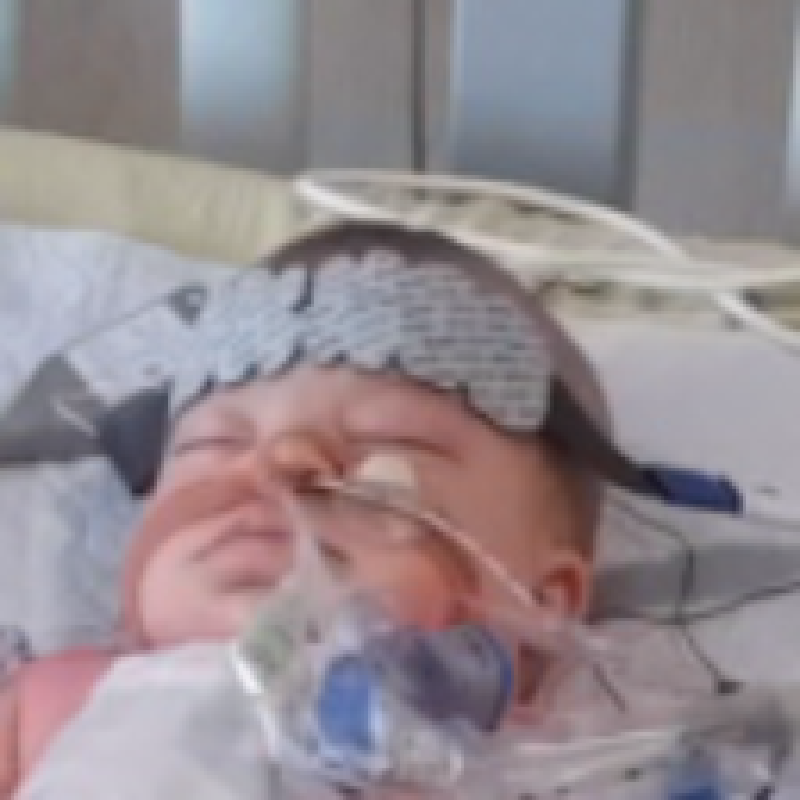}\\[-1pt]
  \includegraphics[width=\textwidth]{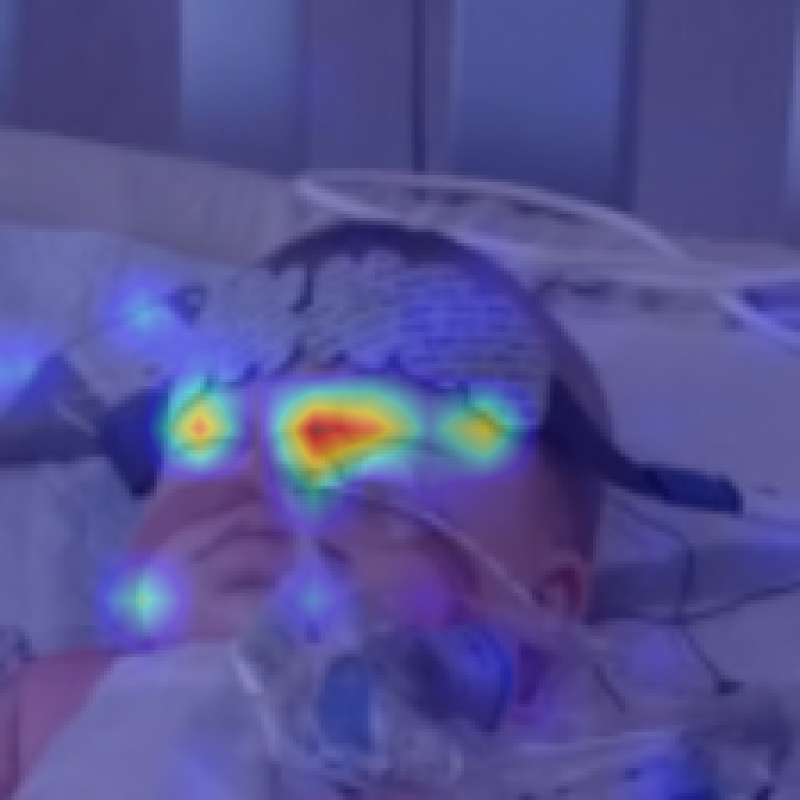}
  \caption*{P4}
\end{subfigure}\hfill
\begin{subfigure}[b]{0.09\textwidth}
  \includegraphics[width=\textwidth]{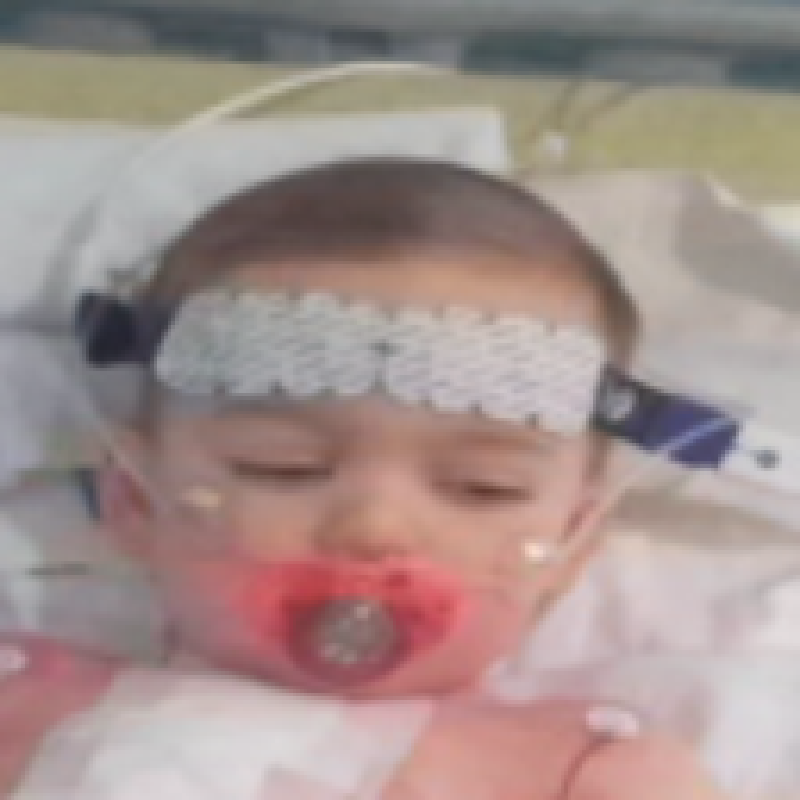}\\[-1pt]
  \includegraphics[width=\textwidth]{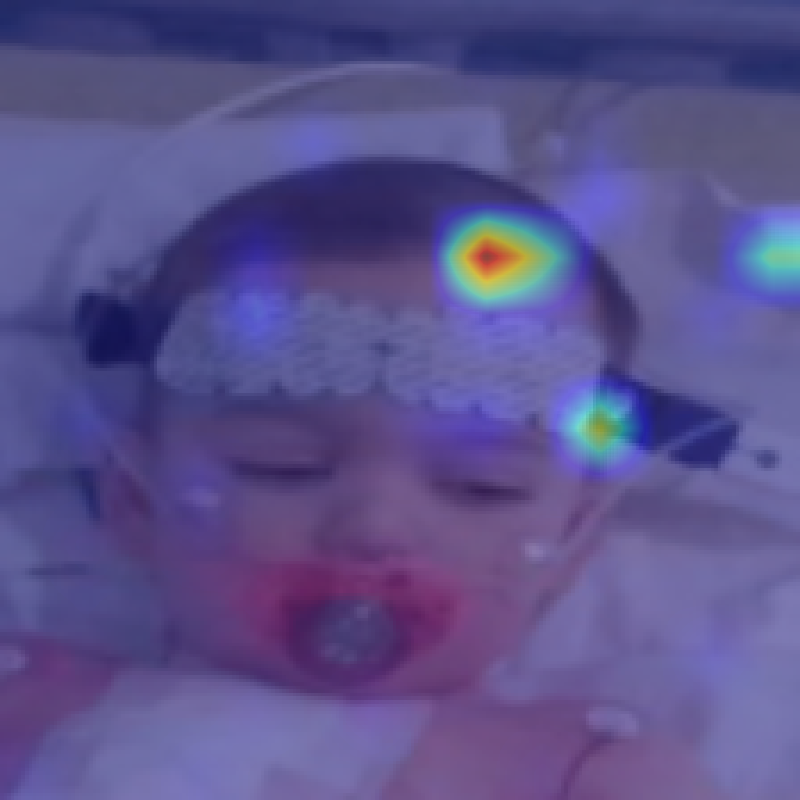}
  \caption*{P5}
\end{subfigure}\hfill
\begin{subfigure}[b]{0.09\textwidth}
  \includegraphics[width=\textwidth]{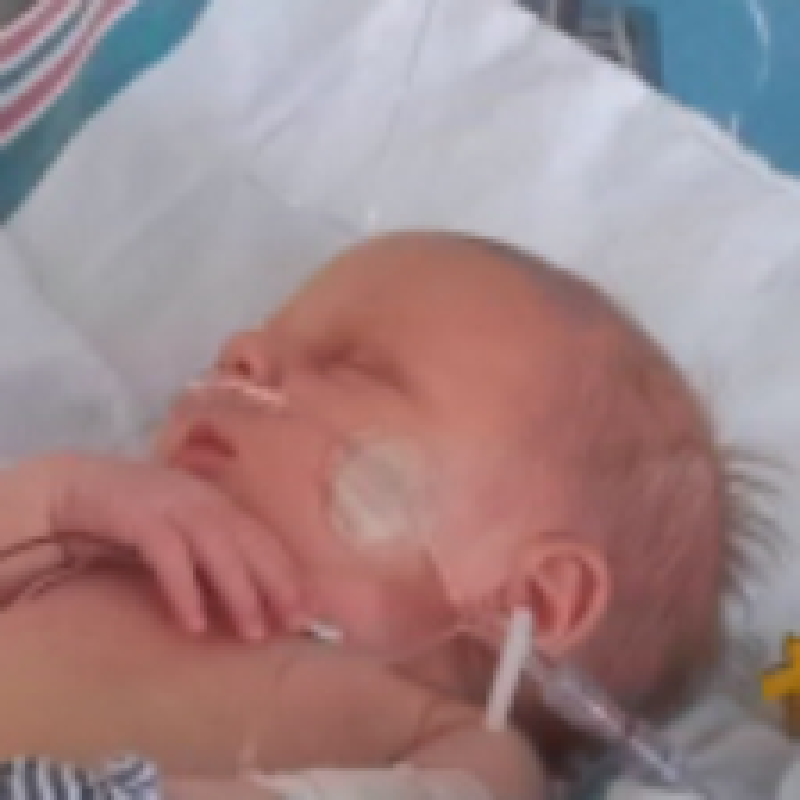}\\[-1pt]
  \includegraphics[width=\textwidth]{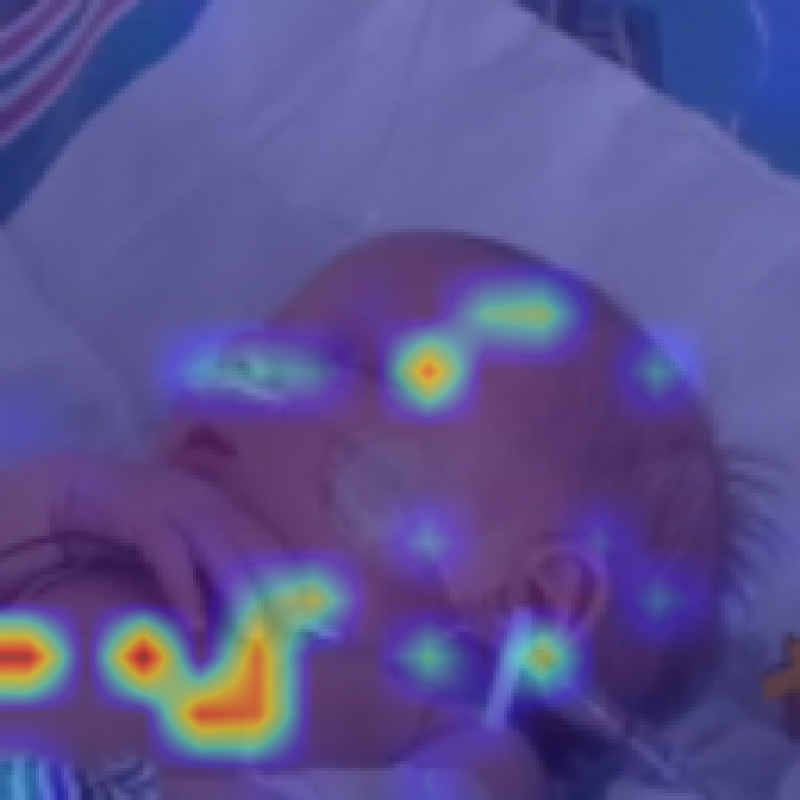}
  \caption*{P6}
\end{subfigure}\hfill
\begin{subfigure}[b]{0.09\textwidth}
  \includegraphics[width=\textwidth]{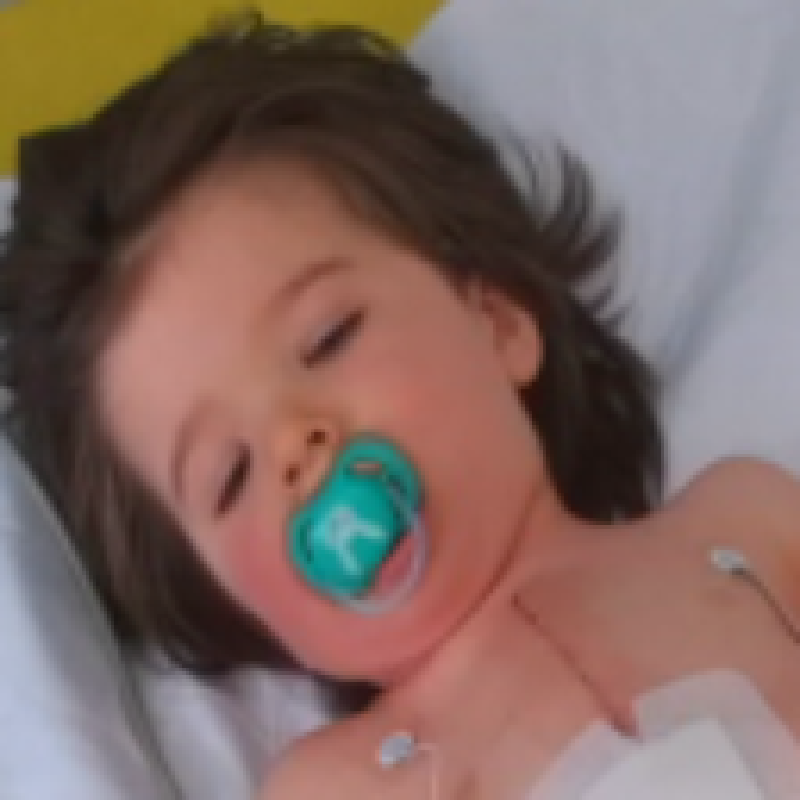}\\[-1pt]
  \includegraphics[width=\textwidth]{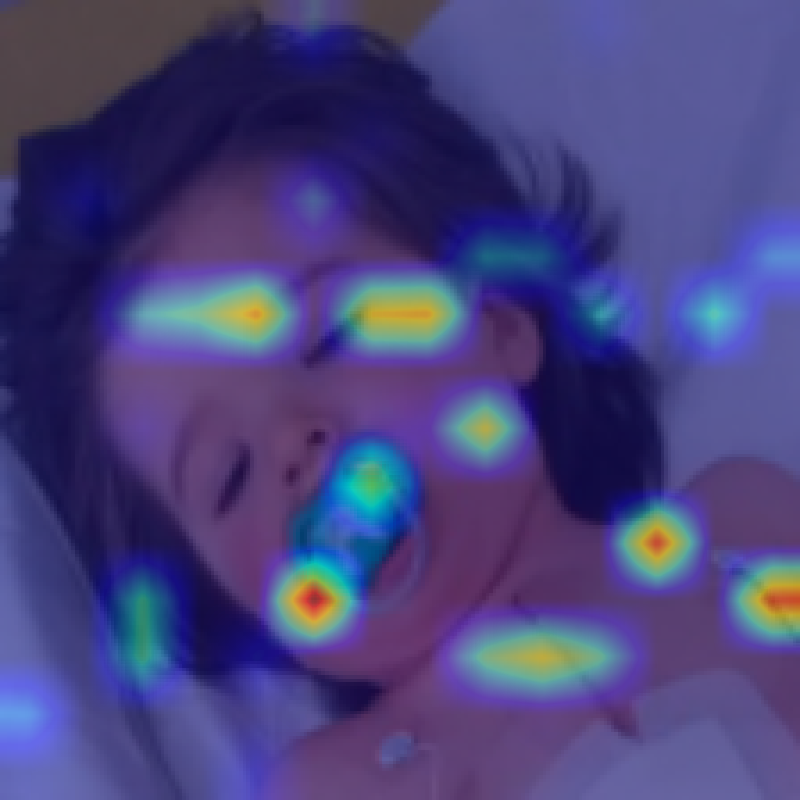}
  \caption*{P7}
\end{subfigure}

\vspace{1mm} 

\begin{subfigure}[b]{0.09\textwidth}
  \includegraphics[width=\textwidth]{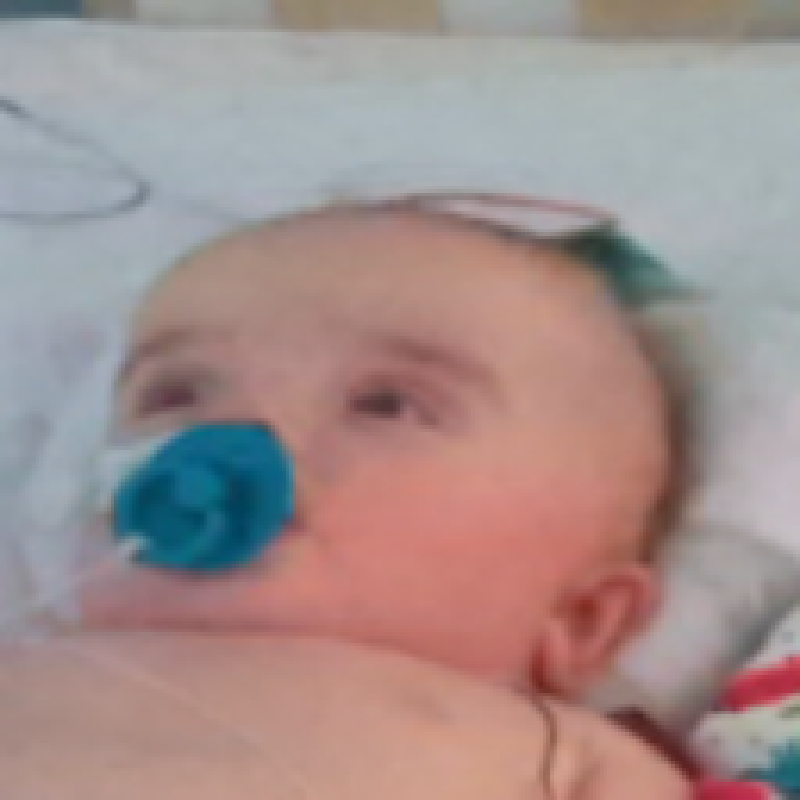}\\[-1pt]
  \includegraphics[width=\textwidth]{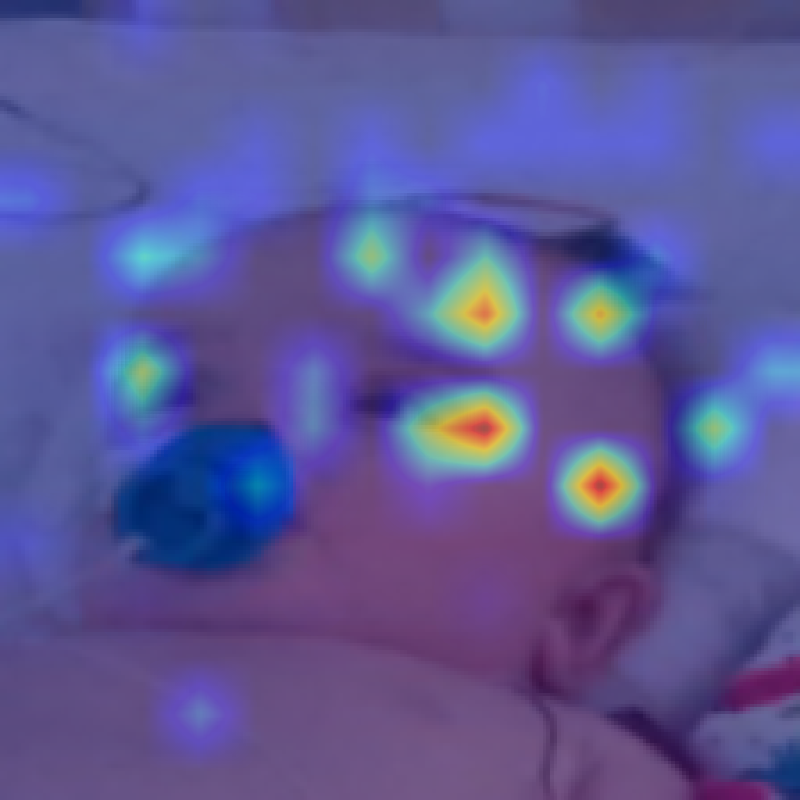}
  \caption*{P8}
\end{subfigure}\hfill
\begin{subfigure}[b]{0.09\textwidth}
  \includegraphics[width=\textwidth]{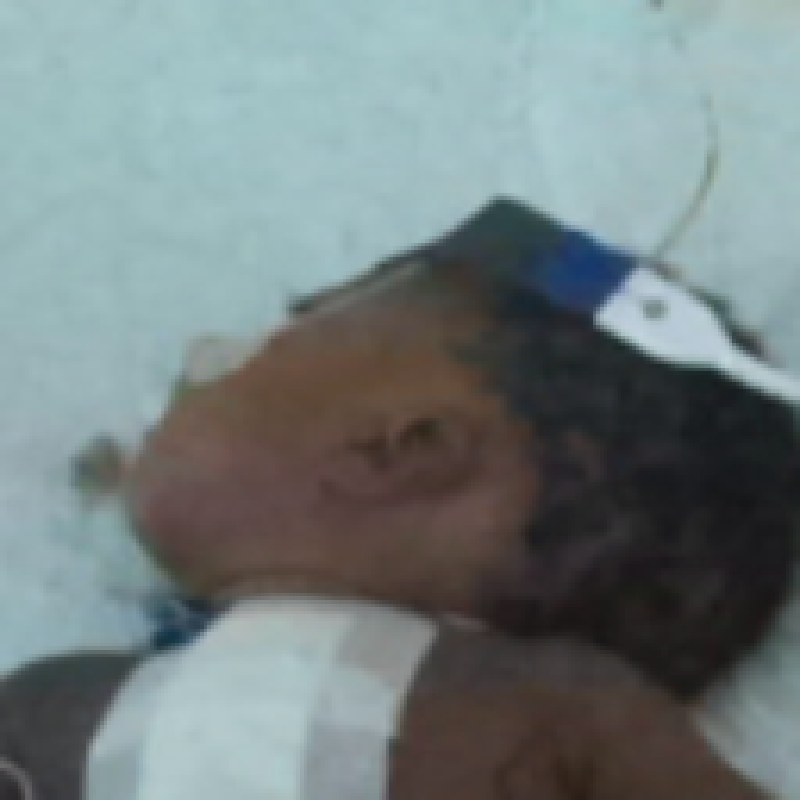}\\[-1pt]
  \includegraphics[width=\textwidth]{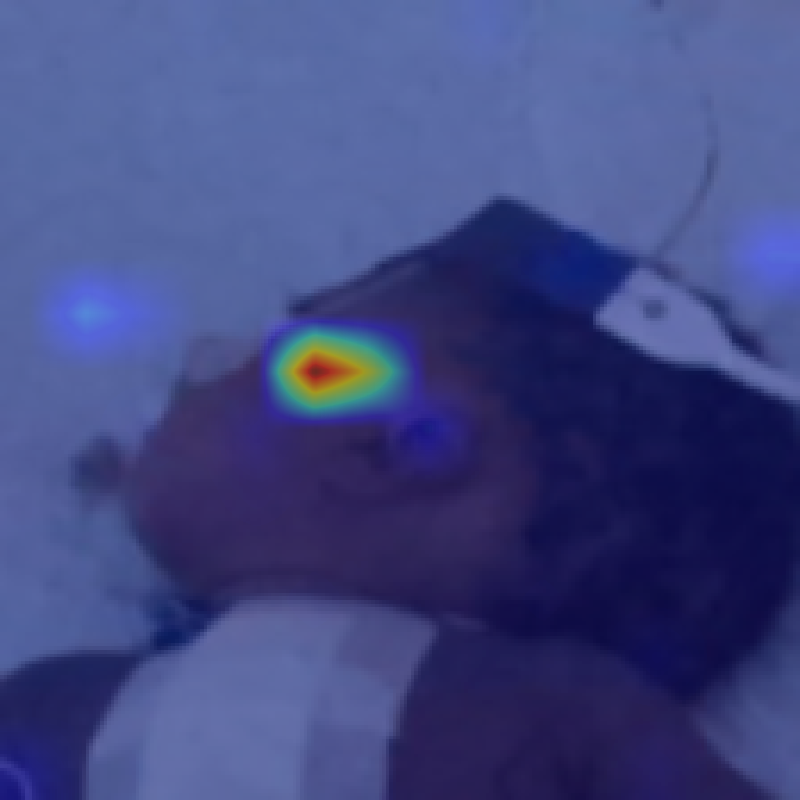}
  \caption*{P9}
\end{subfigure}\hfill
\begin{subfigure}[b]{0.09\textwidth}
  \includegraphics[width=\textwidth]{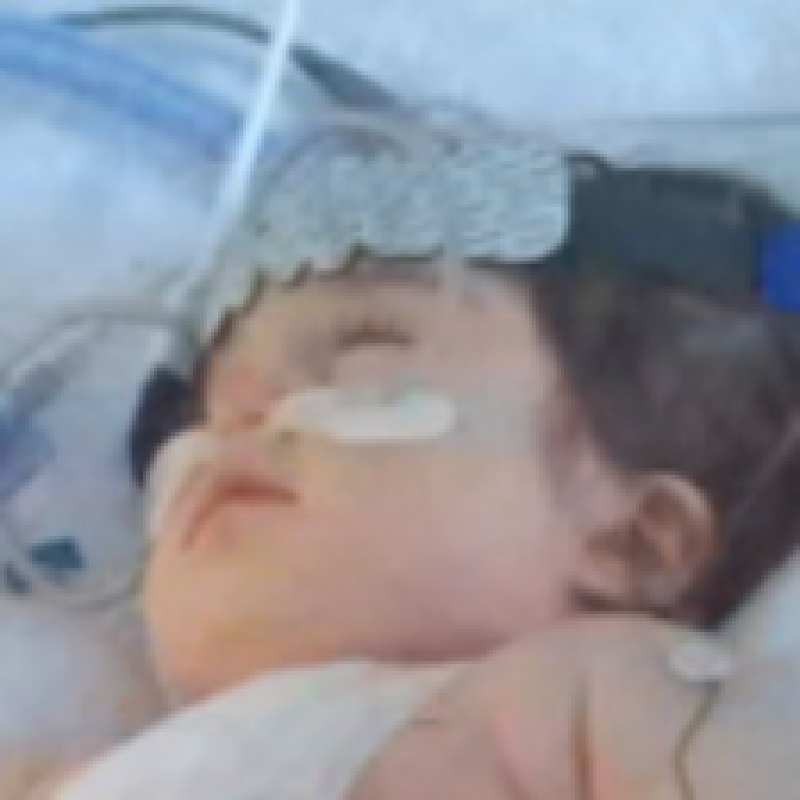}\\[-1pt]
  \includegraphics[width=\textwidth]{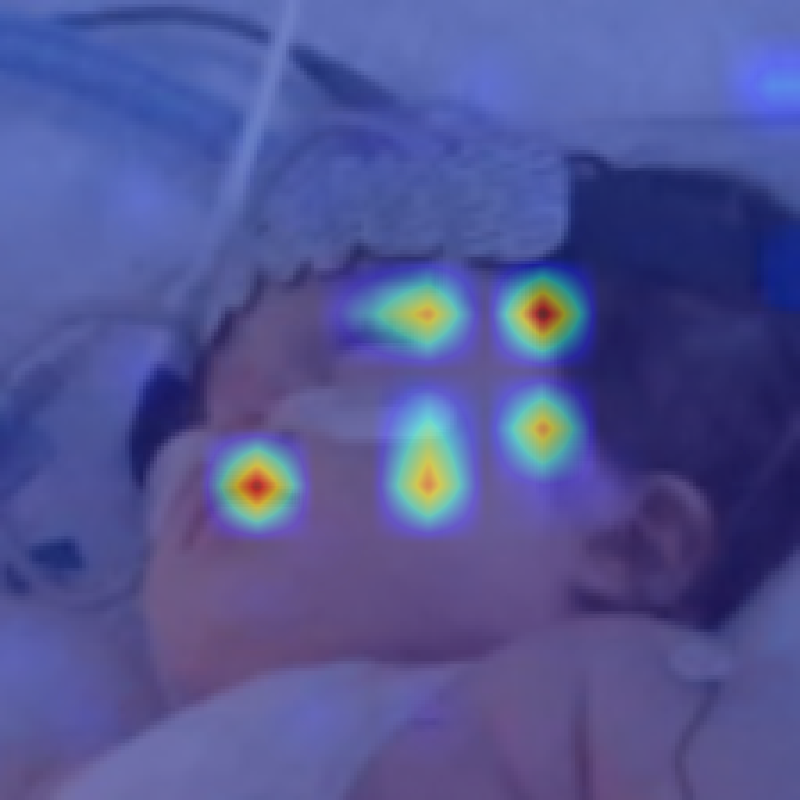}
  \caption*{P10}
\end{subfigure}\hfill
\begin{subfigure}[b]{0.09\textwidth}
  \includegraphics[width=\textwidth]{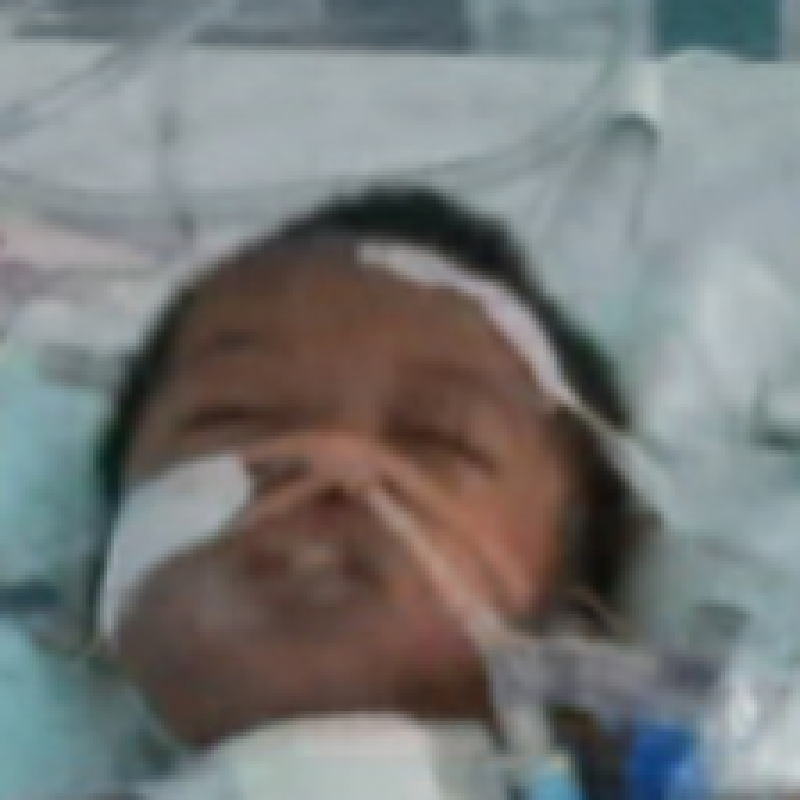}\\[-1pt]
  \includegraphics[width=\textwidth]{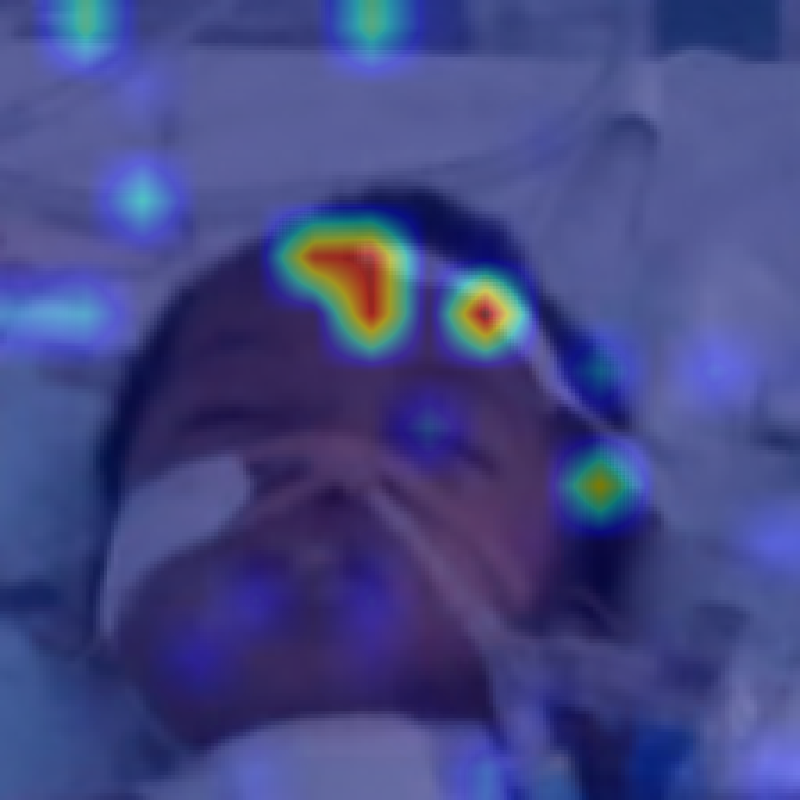}
  \caption*{P11}
\end{subfigure}\hfill
\begin{subfigure}[b]{0.09\textwidth}
  \includegraphics[width=\textwidth]{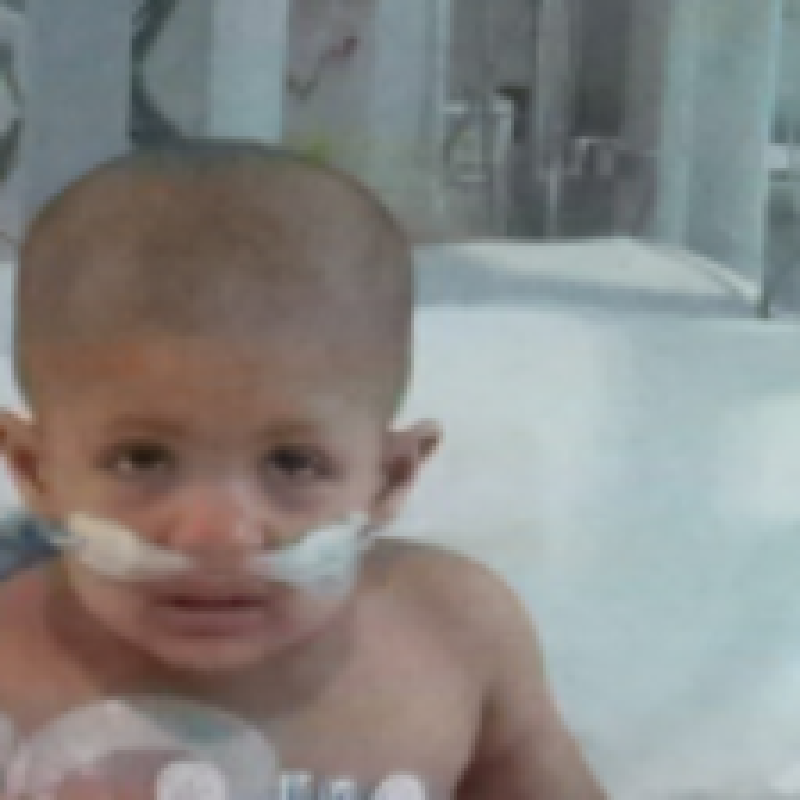}\\[-1pt]
  \includegraphics[width=\textwidth]{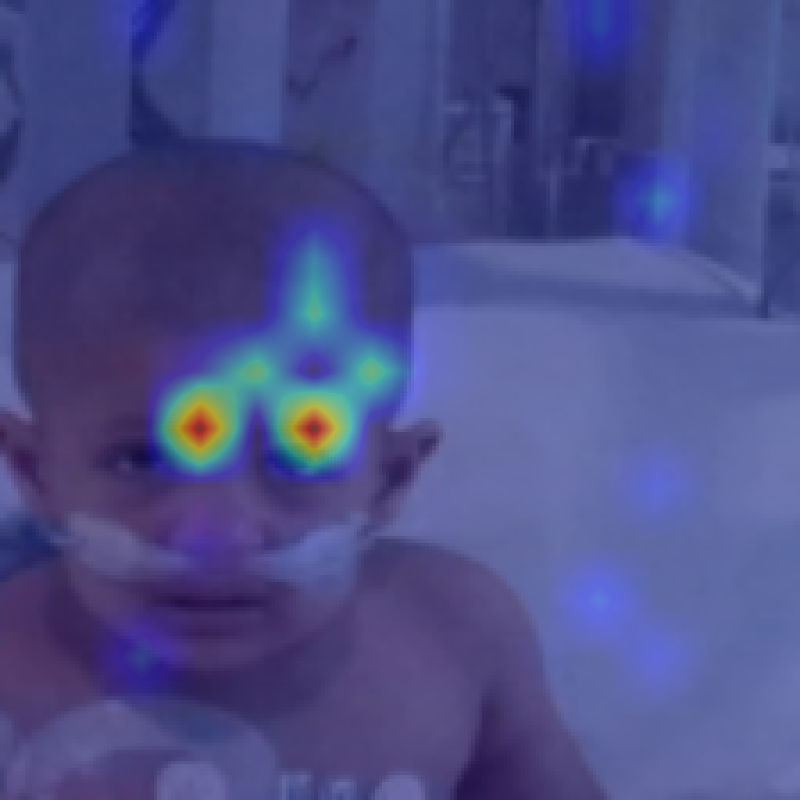}
  \caption*{P12}
\end{subfigure}\hfill
\begin{subfigure}[b]{0.09\textwidth}
  \includegraphics[width=\textwidth]{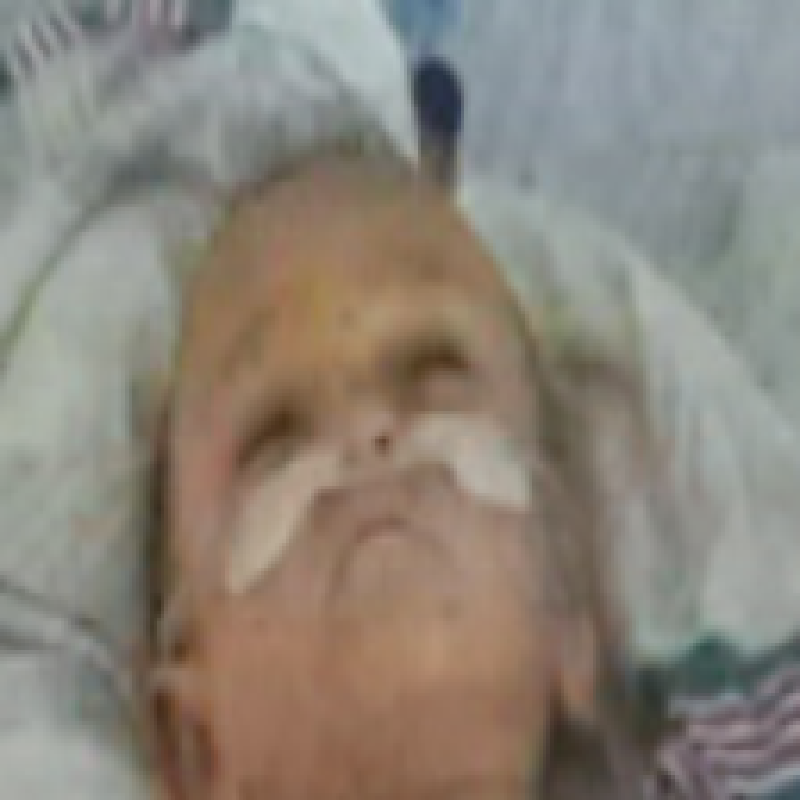}\\[-1pt]
  \includegraphics[width=\textwidth]{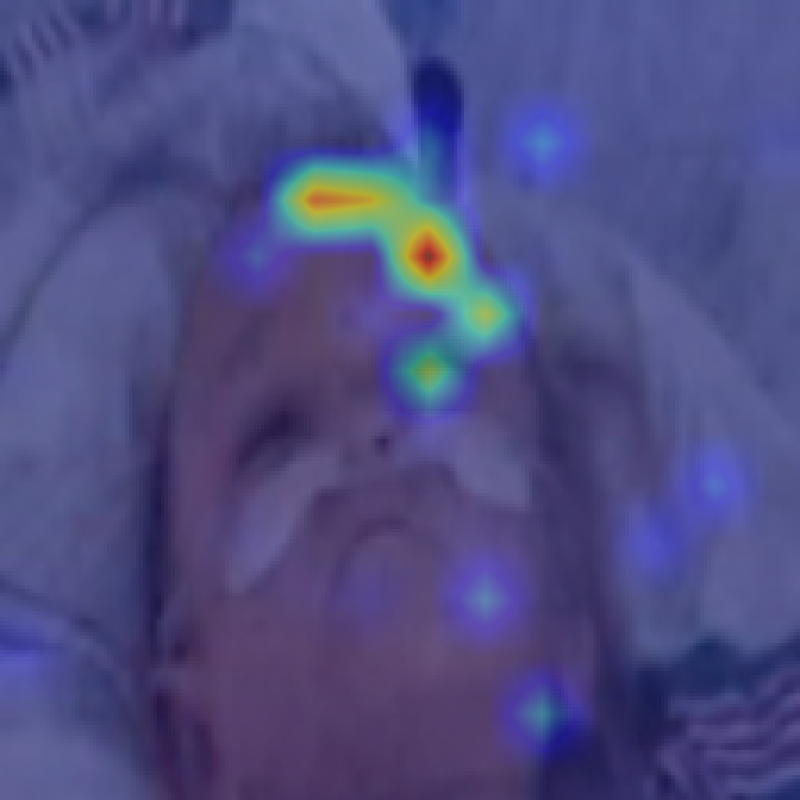}
  \caption*{P13}
\end{subfigure}\hfill
\begin{subfigure}[b]{0.09\textwidth}
  \includegraphics[width=\textwidth]{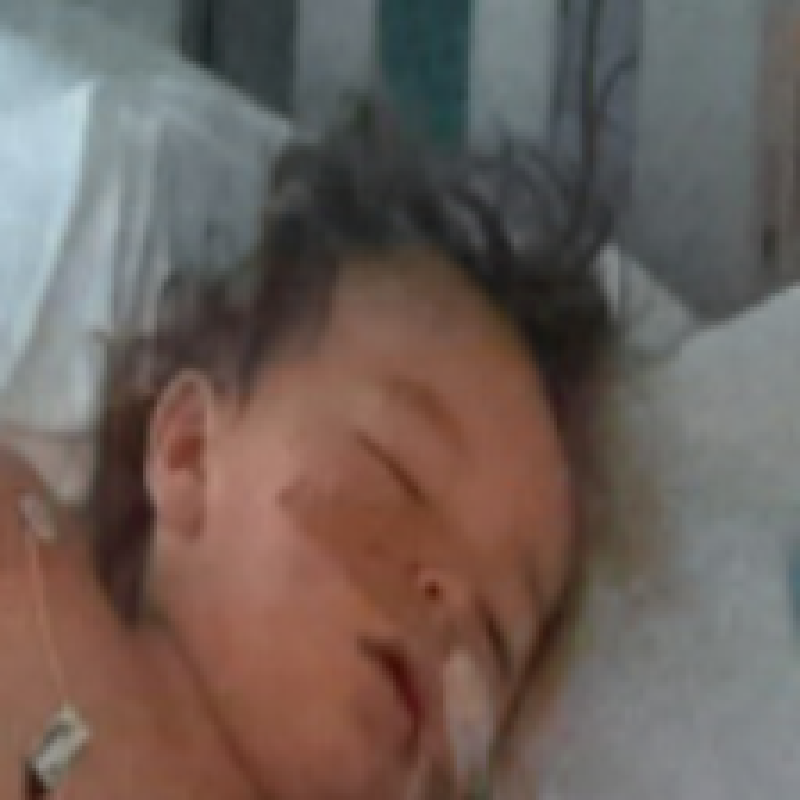}\\[-1pt]
  \includegraphics[width=\textwidth]{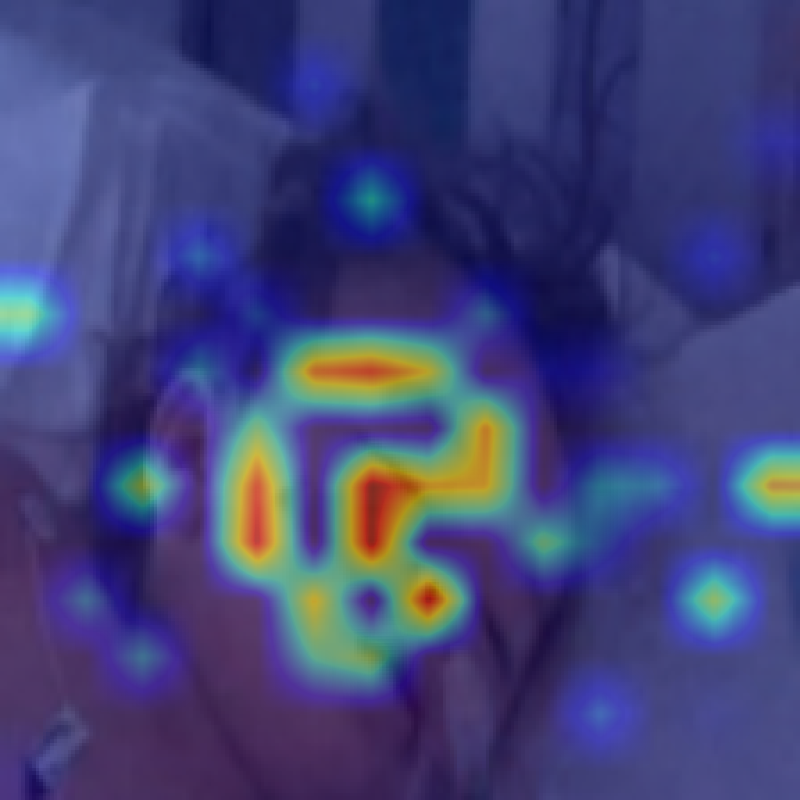}
  \caption*{P14}
\end{subfigure}

\caption{\footnotesize Grad-CAM visualizations from PICU patients, showing the network’s focus on physiologically relevant areas such as the forehead, periorbital region, and thorax. All parents provided written consent for publication.}
\label{fig:picu_gradcam_grid}
\end{figure}

To evaluate the generalization of the learned attention behavior, we further analyzed Grad-CAM visualizations from real PICU patients recorded under diverse conditions. Figure~\ref{fig:picu_gradcam_grid} shows examples from 14 representative cases with varying levels of occlusion, skin tone, and the presence of medical equipment.

The model autonomously identifies physiologically meaningful areas, including the forehead, periorbital regions, upper thorax, and cheeks, without relying on predefined facial landmarks or anatomical masks. These are the regions where superficial blood vessels are most accessible, and the model consistently attends to them even when part of the face is covered.

The attention maps indicate adaptive spatial behavior in the presence of clinical occlusions such as oxygen masks or medical tubing. When primary pulse-bearing regions are blocked, the model redistributes its attention toward secondary areas that preserve pulsatile information, maintaining signal continuity. This adaptive mechanism shows that the model prioritizes physiological relevance rather than fixed spatial locations.

The learned attention strategy remains consistent across patient age groups and clinical conditions. Similar spatial focus is observed in neonates (e.g., P6) and older children (e.g., P3, P11), as well as across skin tones (e.g., P1, P4 vs. P9, P12). This invariance supports the physiological basis of the attention mechanism and suggests that region selection is driven by underlying pulse characteristics.

Compared to the laboratory results in Figure~\ref{fig:grad_cam_vis}, the PICU attention maps show sharper spatial localization and more targeted region prioritization. This improvement likely results from adversarial masking during pretraining, which encourages reliance on the most informative pulse-bearing regions. When these areas are occluded, the model broadens its focus to secondary visible regions, indicating a compensatory mechanism that maintains signal integrity.

This automatic discovery of relevant regions removes the need for separate region-of-interest detection steps, reducing preprocessing complexity and improving robustness. The model learns to segment and weight physiologically meaningful areas end-to-end, enabling practical deployment in clinical monitoring environments.

\subsection{Progressive Pretraining Stages}

     

\begin{table}[htbp]
  \centering
  \footnotesize 
  \renewcommand{\arraystretch}{1.1} 
  \setlength{\tabcolsep}{3pt}      

  \begin{minipage}[t]{0.42\textwidth}
    \vspace{0pt} 
    \centering
    \caption{Ablation of Learning Stages.}
    \label{tab:ablation_curriculum}
    \begin{tabular}{lcccc}
      \toprule
      \textbf{Config.} & \textbf{MAE} & \textbf{RMSE} & \textbf{R} \\
      \midrule
      Direct Sup.      & 18.2 & 22.4 & 0.41 \\
      Stage 1          & 10.3 & 13.6 & 0.64 \\
      Stage 1+2        & 6.8  & 9.2  & 0.78 \\
      \textbf{Full Stage 1-3} & \textbf{3.2} & \textbf{5.4} & \textbf{0.91} \\
      \bottomrule
    \end{tabular}
  \end{minipage}
  \hfill
  \begin{minipage}[t]{0.54\textwidth}
    \vspace{0pt} 
    \centering
    \caption{Ablation of Masking Strategies.}
    \label{tab:ablation_masking}
    \begin{tabular}{lcccc}
      \toprule
      \textbf{Strategy} & \textbf{MAE} & \textbf{RMSE} & \textbf{R} & \textbf{Time} \\
      \midrule
      No Masking (AE)   & 8.4 & 11.2 & 0.70 & 1.0$\times$ \\
      Random (75\%)     & 5.6 & 7.9  & 0.83 & 1.2$\times$ \\
      Random (90\%)     & 6.1 & 8.4  & 0.80 & 1.2$\times$ \\
      Tube (VideoMAE)   & 5.2 & 7.5  & 0.85 & 1.3$\times$ \\
      Adaptive (no PG)  & 4.8 & 7.1  & 0.86 & 1.5$\times$ \\
      \textbf{Full AMN} & \textbf{3.2} & \textbf{5.4} & \textbf{0.91} & \textbf{1.8$\times$} \\
      \bottomrule
    \end{tabular}
  \end{minipage}
\end{table}

Table~\ref{tab:ablation_curriculum} quantifies the cumulative benefits of our three-stage curriculum learning strategy. Compared to direct supervised training on 160 PICU patients, each stage contributes measurable performance improvements that accumulate across the pipeline.

Stage 1 establishes core physiological representations using self-supervised pretraining on clean public datasets. This alone reduces MAE from 18.2 bpm to 10.3 bpm and raises the correlation from 0.41 to 0.64, confirming that exposure to controlled conditions enhances the model’s ability to extract generalizable pulse features.

Stage 2 introduces synthetic clinical artifacts through our hospital occlusion simulator, preparing the model for real-world challenges. Critically, these synthetic occlusions force the model to search beyond easily accessible regions and discover alternative physiological pathways for pulse detection. By dynamically masking different facial regions during training, the model is compelled to explore multiple pathways for rPPG estimation rather than relying on fixed spatial cues. This encourages learning signal-rich zones based on their physiological utility, not on explicit ROI supervision. Consequently, the model becomes more robust to occlusions and better aligned with the anatomical variability encountered in clinical videos. As a result, MAE further decreases to 6.8 bpm and correlation rises to 0.78.

Stage 3 bridges the domain gap by adapting the model to unlabeled PICU videos. Fine-tuning on this real-world data achieves a final MAE of 3.2 bpm and a correlation of 0.91. The complete pipeline yields significant gains in both accuracy and robustness by allowing the model to learn spatial priors from data rather than from explicit supervision.

Overall, the results validate our hypothesis: progressive complexity scheduling, coupled with spatially unconstrained learning, enables effective adaptation to noisy, occluded, and clinically diverse conditions, even under limited ground truth supervision.

\subsection{Adaptive Masking Strategy}
\begin{figure}[htbp]
    \centering
    \captionsetup[subfigure]{font=scriptsize, labelfont=scriptsize}

    \begin{subfigure}[b]{0.45\textwidth}
        \centering
        \includegraphics[width=\textwidth]{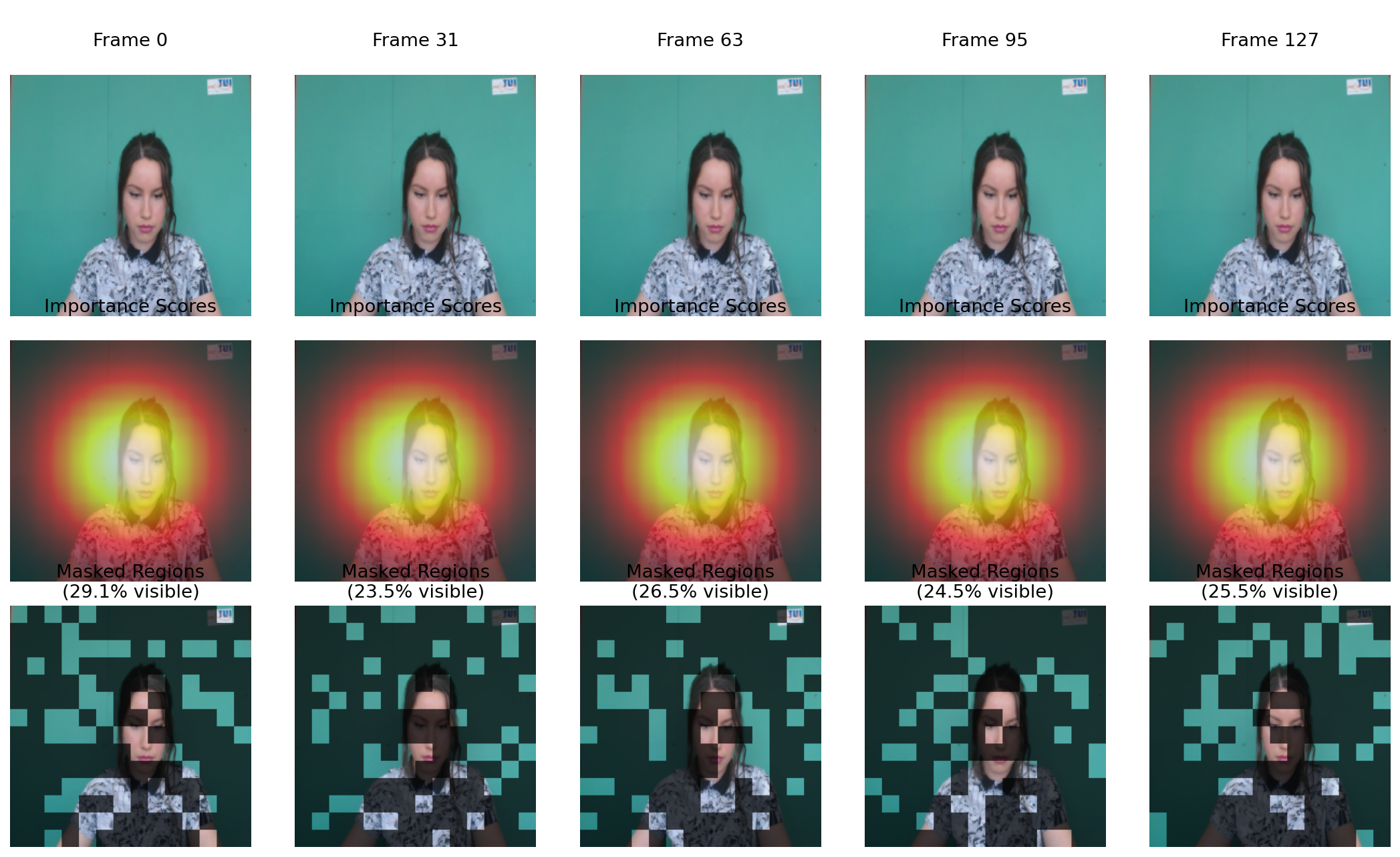}
        \caption{Temporal progression}
        \label{fig:amn_temporal}
    \end{subfigure}
    \hfill
    \begin{subfigure}[b]{0.45\textwidth}
        \centering
        \includegraphics[width=\textwidth, keepaspectratio]{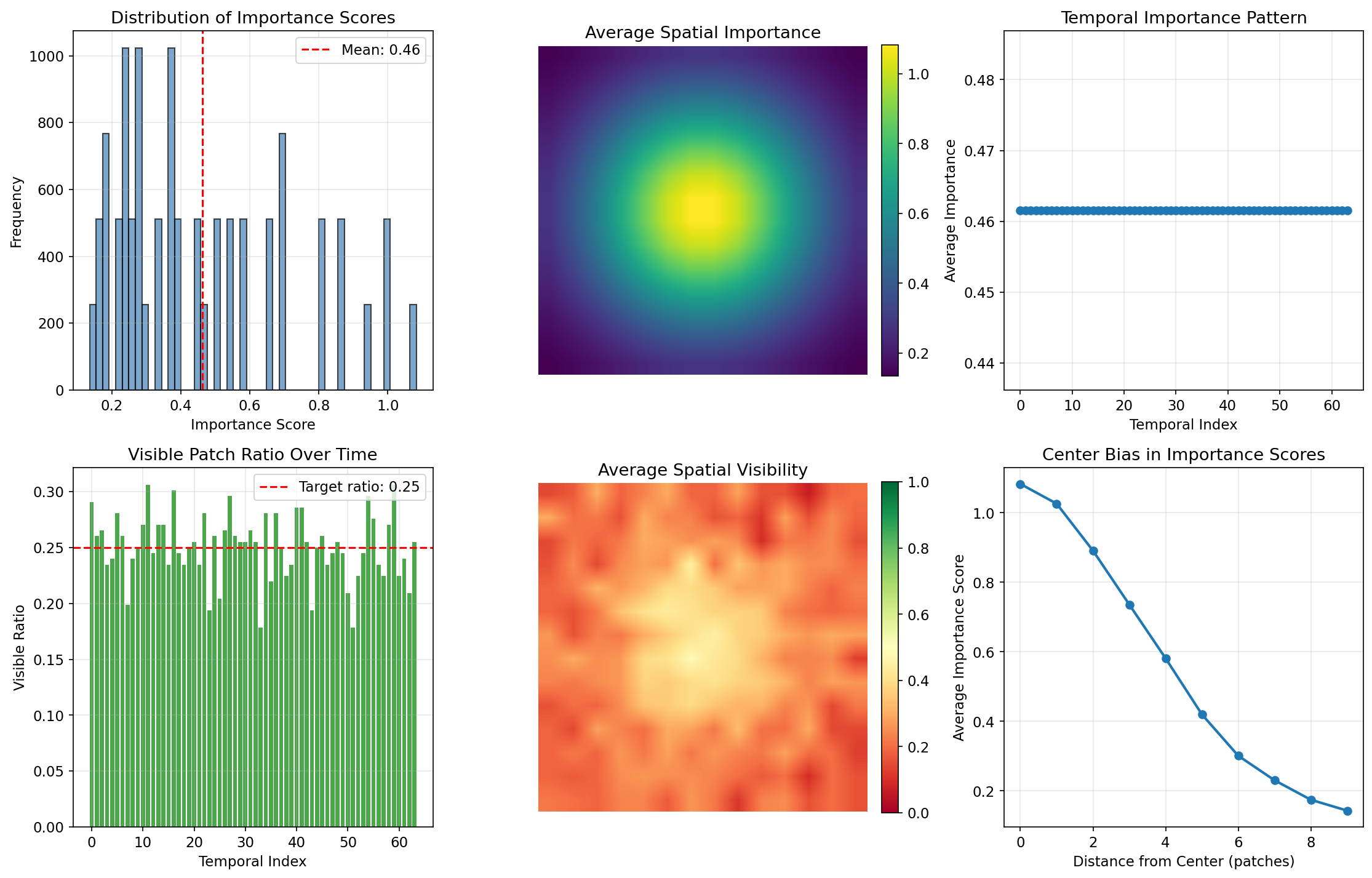}
        \caption{Policy analysis}
        \label{fig:amn_statistics}
    \end{subfigure}

    \caption{\footnotesize Qualitative and quantitative analysis of the Adaptive Masking Network. The AMN consistently identifies salient physiological regions (a) while maintaining a highly selective, bimodal importance distribution (b).}
    \label{fig:amn_comprehensive_analysis}
\end{figure}

Figure~\ref{fig:amn_temporal} presents the temporal evolution of our adaptive masking strategy over 128 frames. The middle row shows that the AMN consistently assigns high importance scores to key pulse-bearing regions such as the forehead and upper cheeks. Despite this spatial focus, the bottom row demonstrates that the AMN generates temporally diverse masking patterns while preserving the target 25\% visibility ratio. This behavior prevents the student from memorizing fixed occlusion patterns and encourages the extraction of robust, redundant features from various facial regions.


Figure~\ref{fig:amn_statistics} quantifies the statistical properties of the learned masking policy. The predicted importance scores follow a bimodal distribution, with distinct peaks at approximately 0.20 and 0.75, corresponding to non-informative and physiologically relevant regions, respectively. The spatial average map exhibits a strong center bias with exponential attenuation toward the periphery, highlighting the AMN’s spatial selectivity. Temporally, the patch visibility remains stable around 25\% ±2\%, confirming that the model does not collapse into fixed or static configurations.
These results suggest that the AMN has converged to a stable yet dynamic masking policy that balances spatial selectivity with temporal variation. By enforcing reconstruction from different patch subsets, the model is pushed to learn redundant, generalizable representations of pulse-rich facial regions, which directly supports robustness under real-world occlusion.

Table~\ref{tab:ablation_masking} evaluates the impact of different masking strategies. Our full AMN approach achieves the lowest MAE (3.2 bpm) and highest correlation (0.91), significantly outperforming baseline methods. Compared to tube-based masking (5.2 bpm), adaptive masking yields a 38.5\% improvement in MAE. Random masking strategies are less effective, and removing the policy gradient degrades performance to 4.8 bpm. Although the full AMN increases training time to 1.8, its gains in accuracy and robustness justify the computational cost for clinical deployment. These results support the utility of adversarial masking to promote learning of transferable features across occluded and unconstrained facial inputs.

     

\subsection{Loss Function Components}

Table~\ref{tab:ablation_loss} summarizes the impact of each loss component and their combinations. Adding the correlation loss to the pixel-wise MSE reduces the MAE from 7.3 to 5.1~bpm, corresponding to a 30.1\% improvement. When used alone, the distillation objective achieves an MAE of 4.6~bpm, representing a 37.0\% reduction compared to MSE. The combination of both objectives yields the lowest MAE of 3.2~bpm, resulting in a 56.2\% improvement over MSE, 37.3\% over MSE+Corr, and 30.4\% over MSE+Distill. These results confirm that pixel-level fidelity and physiological guidance act as complementary supervisory signals, improving both estimation accuracy and temporal consistency.

\begin{table}[htbp]
  \centering
     
  \scriptsize 
  \caption{Ablation of Loss Components.}
  \label{tab:ablation_loss}
  \begin{tabular}{lccc}
    \toprule
    \textbf{Loss Configuration} & \textbf{MAE (bpm)} & \textbf{RMSE (bpm)} & \textbf{R} \\
    \midrule
    MSE only ($L_{\text{pixel}}$) & 7.3 & 9.8 & 0.73 \\
    MSE + Correlation ($L_{\text{recon}}$) & 5.1 & 7.3 & 0.84 \\
    MSE + Distillation & 4.6 & 6.8 & 0.87 \\
    \textbf{Full ($L_{\text{recon}} + L_{\text{distill}}$)} & \textbf{3.2} & \textbf{5.4} & \textbf{0.91} \\
    \bottomrule
  \end{tabular}%
\end{table}

\subsection{Architecture Components}

In critical care environments such as the PICU, real-time inference and hardware constraints demand efficient and deployable models. VisionMamba was selected for its low-latency and lightweight design, offering a favorable trade-off between speed, memory, and accuracy. Table~\ref{tab:ablation_architecture} presents a comparative analysis of computational metrics across several backbone architectures.

VisionMamba achieves the fastest inference time at 85~ms, outperforming ViT-B by 7~ms and reducing latency by 17\%. It also requires the fewest operations (15.5~GFLOPs) and consumes the least memory (1.1~GB), while maintaining a compact model size of 50.2M parameters. These characteristics support its integration into clinical pipelines, where uninterrupted, low-overhead processing is essential for continuous rPPG estimation without interfering with patient care.

\begin{table}[htbp]
  \centering
  \scriptsize  
  \caption{Comparison of computational efficiency between encoder backbones}
  \label{tab:ablation_architecture}
  \begin{tabular}{lcccc}
    \toprule
    \textbf{Architecture} & \textbf{Params (M)} & \textbf{FLOPs (G)} & \textbf{Memory (GB)} & \textbf{Inference Time (ms)} \\
    \midrule
        ViT-B (Baseline) & 86.4 & 16.1 & 7.3 & 92 \\
        TimeSformer      & 121.3 & 379.8 & 8.6 & 300 \\
        VideoMAE         & 86.2 & 101.85 & 7.9 & 108 \\
    \textbf{VisionMamba (Ours)} & \textbf{50.2} & \textbf{15.5} & \textbf{1.1} & \textbf{85} \\
    \bottomrule
  \end{tabular}
\end{table}

\section{Conclusion}
Continuous and unobtrusive monitoring of vital signs in the PICU requires models that are accurate, lightweight, and robust to real-world clinical conditions. This work presents a self-supervised rPPG estimation framework tailored to these demands. Based on a VisionMamba backbone, the proposed system integrates an Adaptive Masking Network to enforce spatial-temporal relevance and a physiological distillation head to guide signal consistency. 
Evaluated on realistic PICU videos, the method achieves 3.2 bpm MAE with strong temporal correlation, while operating within clinical latency and resource constraints (85 ms inference time, 15.5 GFLOPs, 1.1 GB memory). The masking strategy provides spatial selectivity and temporal diversity under occlusion, and the distillation mechanism enhances physiological fidelity across sequences. These design choices allow the model to generalize across challenging lighting conditions and partial obstructions common in bedside recordings.
Future work will focus on expanding the demographic diversity of the training data, conducting multi-institutional validation, and extending the approach to respiration monitoring. Additional efforts will address real-time adaptation and integration with embedded systems to enable secure, continuous inference in the clinical workflow.

\end{document}